\documentclass[twoside,leqno,twocolumn]{article}

\usepackage[letterpaper]{geometry}

\usepackage{ltexpprt}
\usepackage{amsmath,amssymb,amsfonts}
\usepackage{graphicx}
\usepackage{textcomp}
\usepackage{xcolor}
\usepackage{hyperref}
\usepackage{stackengine}
\usepackage{subfig}
\usepackage{hyperref}
\usepackage{balance}
\usepackage{algorithmic,algorithm}

\begin{document}

\newcommand\relatedversion{}

\title{\Large AVATAR: Adversarial Autoencoders with Autoregressive Refinement for Time Series Generation}
\author{ MohammadReza EskandariNasab\thanks{Utah State University. Emails: \{reza.eskandarinasab, s.hamdi, soukaina.boubrahimi\}@usu.edu.} \thanks{Corresponding author (ORCID: 0009-0004-0697-3716).}
\and Shah Muhammad Hamdi\footnotemark[1]
\and Soukaina Filali Boubrahimi\footnotemark[1]
}
\date{}

\maketitle


\fancyfoot[R]{\scriptsize{Copyright \textcopyright\ 2025 by SIAM\\
Unauthorized reproduction of this article is prohibited}}





\begin{abstract} \small\baselineskip=9pt 
Data augmentation can significantly enhance the performance of machine learning tasks by addressing data scarcity and improving generalization. However, generating time series data presents unique challenges. A model must not only learn a probability distribution that reflects the real data distribution but also capture the conditional distribution at each time step to preserve the inherent temporal dependencies. To address these challenges, we introduce AVATAR, a framework that combines Adversarial Autoencoders (AAE) with Autoregressive Learning to achieve both objectives. Specifically, our technique integrates the autoencoder with a supervisor and introduces a novel supervised loss to assist the decoder in learning the temporal dynamics of time series data. Additionally, we propose another innovative loss function, termed distribution loss, to guide the encoder in more efficiently aligning the aggregated posterior of the autoencoder’s latent representation with a prior Gaussian distribution. Furthermore, our framework employs a joint training mechanism to simultaneously train all networks using a combined loss, thereby fulfilling the dual objectives of time series generation. We evaluate our technique across a variety of time series datasets with diverse characteristics. Our experiments demonstrate significant improvements in both the quality and practical utility of the generated data, as assessed by various qualitative and quantitative metrics. \end{abstract}\\


\section{Introduction.}
Data scarcity poses a significant challenge in many real-world, machine learning-based prediction tasks \cite{Ahmadzadeh2021}. This issue can be particularly pronounced for certain classes within a dataset, especially when dealing with rare events such as major solar flares \cite{EskandariNasab2024, sfcontrastive-reza, sfcontrastive-onur}, or patients with specific medical conditions, such as cancer. Additionally, data scarcity may arise from privacy concerns \cite{Abouelmehdi2017, Cascalheira2024, santosh2024}, especially in healthcare, or due to noisy and complex data collection environments \cite{Vincent2017}. Beyond individual classes, data scarcity can affect entire datasets, as deep learning models are notoriously data-hungry \cite{Mikolajczyk2018}, often requiring vast amounts of data to achieve optimal performance. To address these challenges, data augmentation techniques are commonly employed to artificially increase the size of datasets and facilitate the training of machine learning and deep learning models \cite{thesis24}. This approach has found widespread use in fields such as computer vision \cite{Voulodimos2018, aria2022}, time series analysis \cite{Li2023, obahri2025, AlshammariTF2024}, and signal processing \cite{ESKANDARINASAB2024GRUCNN}. However, the quality of the augmented data is critical, especially in time series analysis, where models must capture not only the distribution of features at individual time points but also the complex interactions between these points over time. For instance, in multivariate sequential data \(x_{1:T} = (x_1, \ldots, x_T)\), an effective model must accurately capture the conditional distribution \(p(x_t \mid x_{1:t-1})\), which governs temporal transitions. Failure to do so leads to poor model performance, underscoring the importance of preserving temporal dynamics in time series data.

Considerable research has been directed towards enhancing the temporal behavior of autoregressive models used in sequence forecasting. The primary aim is to reduce the effect of sampling errors by making targeted adjustments during training, thereby improving the modeling of conditional distributions \cite{ar1, ar2, ar3}. Autoregressive models represent the sequence distribution as a product of conditional probabilities \(\prod_{t} p(x_t \mid x_{1:t-1})\), which makes them well-suited for forecasting due to their deterministic nature. However, these models are not truly generative, as they do not rely on external inputs to generate new data sequences.

In contrast, studies applying generative adversarial networks (GANs) \cite{gan} to sequential data generally employ recurrent neural networks (RNNs) for both the generator and discriminator, aiming to optimize an adversarial objective \cite{gan1, gan2}. While this approach is conceptually simple, the adversarial objective is crafted to capture the joint distribution \(p(x_{1:T})\), which doesn't inherently address the autoregressive nature of the data. This can lead to issues, as summing traditional GAN losses over fixed-length vectors might not effectively model the stepwise dependencies characteristic of sequence data. Moreover, GAN-based methods face several well-known challenges, including mode collapse, instability, and non-convergence, which can complicate the training process.

Adversarial Autoencoders (AAEs) \cite{aae} offer several advantages over GANs. First, AAEs combine the benefits of both autoencoders and adversarial training, allowing for more structured latent space representations. This makes AAEs better at capturing meaningful features from data, leading to more realistic and diverse augmented samples. Unlike GANs, which can suffer from mode collapse (where only a few modes of the data distribution are learned), AAEs are designed to regularize the latent space, helping to maintain a balance across different modes. Additionally, AAEs are typically easier to train compared to GANs, which often require careful tuning to avoid instability in the generator-discriminator dynamic.

Our proposed method, termed \textbf{AVATAR} (\textbf{A}d\textbf{V}ersarial \textbf{A}u\textbf{T}oencoders with \textbf{A}utoregressive \textbf{R}efinement), is designed to leverage multiple research paradigms in order to develop a robust time series generation technique. AVATAR effectively captures the temporal dynamics inherent in time series data while learning the complete data distribution of the dataset, enabling the generation of reliable and realistic time series sequences. Additionally, the model incorporates innovative loss functions that enhance stability and guide the network toward efficiently converging to the global minima. Our key contributions are as follows:

\begin{enumerate}
    \item 
    Combining AAEs with autoregressive learning, which introduces a supervisory network to the probabilistic autoencoder. This supervisory network assists the decoder in learning the conditional distribution of sequences over time. Additionally, we introduce a new supervised loss to achieve enhanced performance in autoregressive learning.
    
    \item 
    We introduce an innovative loss function, referred to as \textit{distribution loss}, which enables the encoder to capture the nuances of the prior Gaussian distribution more efficiently and precisely, resulting in faster convergence and enhanced stability.

    \item 
    Adapting a GRU model with batch normalization as a regularization technique, instead of a standard GRU architecture, which optimizes and enhances the network's performance, resulting in superior generalization.
    
    \item 
    Proposing a joint training approach, where the supervisory network and autoencoder are trained together as an integrated network with a combined loss function. This integrated approach ensures the successful generation of time series data by aligning the learning objectives of both networks.
\end{enumerate}

We illustrate the effectiveness of AVATAR through a series of experiments conducted on a range of real-world and synthetic multivariate time series datasets. Our results demonstrate that AVATAR consistently surpasses existing benchmarks in generating realistic time series data.

\section{Related Work.}

Autoregressive RNNs trained with maximum likelihood estimation frequently encounter significant errors during multi-step predictions \cite{Williams1989}. These errors arise due to the discrepancy between the training phase, conducted in a closed-loop setting (where the model is conditioned on actual data), and the inference phase, which operates in an open-loop manner, relying on previous predictions. To address this issue, approaches such as Professor Forcing, based on adversarial domain adaptation \cite{Pforcing}, have been proposed. This method introduces an additional discriminator designed to distinguish between self-generated hidden states and those derived from teacher-forced inputs \cite{Tforcing}. This alignment of training and prediction improves the consistency of the model’s behavior. However, while these methods attempt to capture stepwise transitions, they remain deterministic and do not involve sampling from a learned probability distribution, an essential requirement for generating synthetic data.

The introduction of GANs \cite{gan} represented a major innovation in data generation. GANs consist of two neural networks, a generator and a discriminator, trained simultaneously in a competitive, zero-sum game framework. Although GANs can effectively generate data by sampling from a learned distribution, they struggle to model sequential dependencies, which are crucial for time series data \cite{Brophy2023}. The adversarial feedback from the discriminator alone is often insufficient for the generator to accurately learn and replicate the temporal patterns necessary for time series generation.

Several adaptations of the GAN framework have been developed specifically for time series data \cite{seriesgan}. One of the earliest models, C-RNN-GAN \cite{gan1}, applied GANs to sequential data using LSTM networks for both the generator and the discriminator. This model generates sequences recurrently, starting with a noise vector and using the output from the previous time step to generate the next. RCGAN \cite{gan2} improved upon this approach by eliminating the reliance on previous outputs and incorporating conditional inputs for greater control and flexibility in data generation. Nevertheless, both models are limited by their exclusive reliance on adversarial feedback, which does not fully capture the complex temporal dynamics present in time series data. TimeGAN \cite{timegan} represents an advanced approach for time series generation, incorporating elements of both unsupervised learning (GANs) and supervised training. TimeGAN optimizes both a GAN and an embedding space through a combination of adversarial and supervised objectives, with the goal of generating data in the embedding space rather than the feature space, which helps the network avoid issues related to non-convergence. Despite its innovative design, TimeGAN faces several challenges. Its reliance on adversarial training within the embedding space degrades the quality of the generated data due to noise introduced in the embedding process. Additionally, TimeGAN suffers from stability issues, often yielding inconsistent results even when identical hyperparameters and iteration counts are used. Furthermore, it struggles to fully learn the underlying probability distribution of the dataset, resulting in generated data that fail to represent the entire population effectively.

AAEs \cite{aae} extend the concept of autoencoders by incorporating adversarial training, aiming to impose a prior distribution on the latent space while maintaining the reconstruction capabilities of standard autoencoders. In contrast to Variational Autoencoders (VAEs) \cite{variational}, which explicitly model the latent space using a variational approach, AAEs employ a discriminator to distinguish between samples drawn from the true prior distribution and those generated by the encoder. This adversarial training forces the encoder to produce latent variables that align with the desired distribution, allowing for more effective sampling and data generation. AAEs make it easier to learn the full distribution of a dataset since the model tries to match the aggregated posterior latent representation with an arbitrary prior distribution, both of which exist in lower dimensions compared to real samples \cite{Creswell2019}. This results in a less noisy, more compact representation that leads to more accurate adversarial training and faster convergence in terms of iteration count. However, AAEs are not designed for time series data and struggle to learn the temporal dependencies in such datasets, leading to the generation of misleading time series data.

Autoregressive models do not rely on sampling from a learned probability distribution to generate data, and AAEs are not capable of effectively capturing the temporal dynamics of time series data, unlike autoregressive models. To address these limitations, AVATAR combines these two research streams to develop a simple yet powerful framework for time series generation. By integrating a teacher forcing-based supervisory network with the autoencoder, introducing novel loss functions, and designing a joint training ecosystem, AVATAR provides a robust framework for generating synthetic time series data across various applications.

\section{Problem Specification.}

Let $\mathcal{X}$ denote the space of the temporal features, and consider $X \in \mathcal{X}$ as random vectors, with each vector attaining particular values represented by $x$. We focus on sequences of temporal data, denoted $\mathbf{X}_{1:T}$, sampled from a joint probability distribution $p$. The sequence length $T$ is itself a random variable integrated into the distribution $p$. The training dataset contains $N$ samples, each indexed by $n \in \{1, \ldots, N\}$, and is expressed as $\mathcal{D} = \{{X}_{n,1:T_n}\}_{n=1}^{N}$. For simplicity, we omit the subscript $n$ unless necessary for clarity.

The objective of AAEs is to leverage the training data $\mathcal{D}$ to estimate a probability density function $\hat{p}(\mathbf{X}_{1:T})$ that approximates the true distribution $p(\mathbf{X}_{1:T})$ as closely as possible. Achieving this is challenging due to the complex nature of the distribution. Additionally, in order to capture the temporal dynamics of data using autoregressive models, our objective is to decompose the joint distribution $p(\mathbf{X}_{1:T})$ in an autoregressive manner as $p(\mathbf{X}_{1:T}) = \prod_t p(X_t|\mathbf{X}_{1:t-1})$. This approach refocuses the task on learning a conditional density function $\hat{p}(X_t|\mathbf{X}_{1:t-1})$ that serves as an approximation of $p(X_t|\mathbf{X}_{1:t-1})$ at each time step $t$. Therefore, we define two key objectives:

\begin{enumerate}
   
  \item
\textbf{Global objective}: This objective involves aligning the joint distribution of entire sequences between real and synthetic data. It is expressed as:
\begin{equation}
    \min_{\hat{p}} D\left( p(\mathbf{X}_{1:T}) \middle\| \hat{p}(\mathbf{X}_{1:T}) \right),
\end{equation}

where $D$ represents an appropriate distance metric between the distributions.
  \item
\textbf{Local objective}: This objective focuses on matching the conditional distributions at each time step for all \(t\), formulated as:
\begin{equation}
    \min_{\hat{p}} D\left( p(X_t|\mathbf{X}_{1:t-1}) \middle\| \hat{p}(X_t|\mathbf{X}_{1:t-1}) \right),
\end{equation}

\end{enumerate}
In the context of data augmentation, the global objective corresponds to minimizing the Jensen-Shannon divergence \cite{divergence1} between real and generated distributions, while the local objective in supervised learning aims to minimize the Kullback-Leibler divergence \cite{divergence2}. Minimizing the global objective assumes access to a perfect discriminator (which is often impractical), whereas minimizing the local objective requires access to ground-truth sequences (which we have in this case). Consequently, this approach combines the adversarial learning goal (focused on global distribution) with a supervised learning objective (targeting conditional distributions).

\section{Proposed Framework: AVATAR.}
\begin{figure}
\centering
\includegraphics[width=0.5\textwidth]{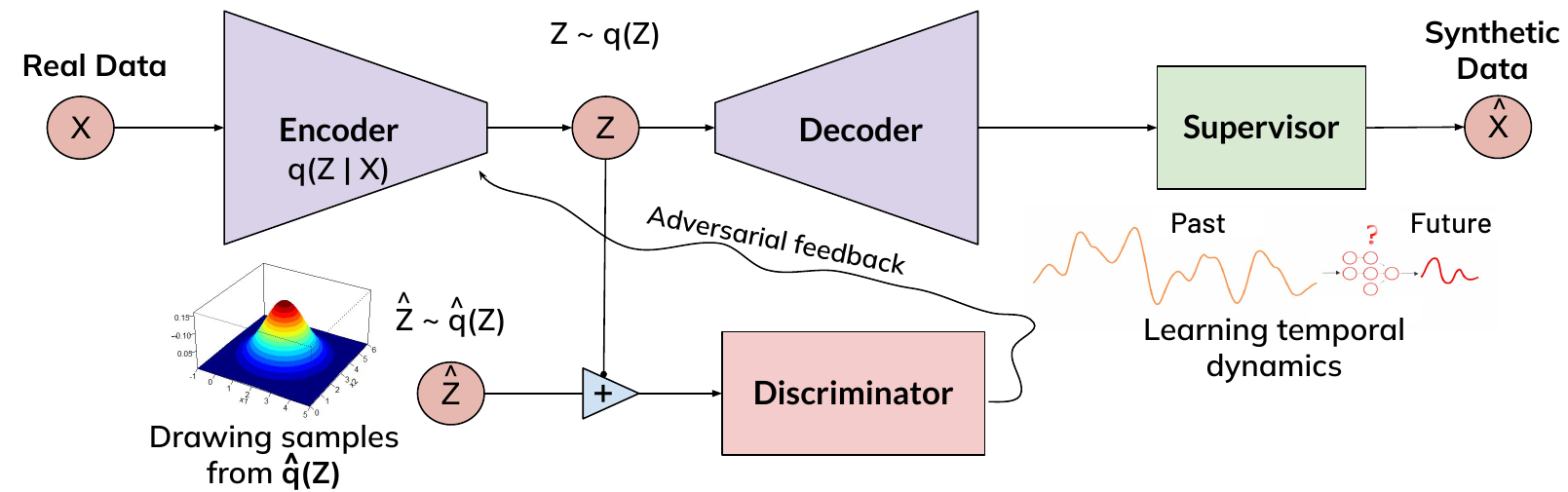}
\caption{The figure illustrates the AVATAR framework designed for time series generation. The autoencoder aims to reconstruct the input \( X \) from its latent representation \( Z \), while the discriminator strives to align the aggregated posterior distribution \( q(z) \) with a prior Gaussian distribution \( \hat{q}(z) \). The supervisor network's role is to assist the decoder in learning the conditional distribution of the data at each time step \( t \).}
\label{fig:avatar}
\end{figure}

Based on Figure~\ref{fig:avatar}, let \( z \) denote the latent representation (hidden code) in the autoencoder with a deep encoder-decoder architecture. We define \( \hat{q}(z) \) as the prior Gaussian distribution we aim to impose on the latent codes, \( q(z|x) \) as the encoding distribution, and \( p(x|z) \) as the decoding distribution. The encoding function \( q(z|x) \) in the autoencoder determines the aggregated posterior distribution of the latent code \( q(z) \).

In AAEs, optimization is achieved by aligning the aggregated posterior \( q(z) \) with a Gaussian prior \( \hat{q}(z) \). Consequently, decoding any area of the prior space produces meaningful synthetic data. To facilitate this alignment, an adversarial network is incorporated at the latent code layer of the autoencoder. The adversarial network's role is to ensure that \( q(z) \) closely matches the prior \( \hat{q}(z) \), while the autoencoder's primary objective is to minimize the reconstruction error. The generator within the adversarial network serves as the encoder of the autoencoder \( q(z|x) \), helping it produce an aggregated posterior that can deceive the discriminator in the adversarial training, making the discriminator believe that the hidden code \( q(z) \) comes from the true prior \( \hat{q}(z) \). Upon the completion of training, the decoder of the autoencoder functions as a generative model, mapping the prior \( \hat{q}(z) \) to the data distribution \( p(x) \).

Although this method is beneficial, it is inadequate on its own for modeling the conditional distribution of time series data at each time step \( t \). To overcome this limitation, as depicted in Figure~\ref{fig:avatar}, we have enhanced the decoder of the AAE by integrating a supervisor network that employs teacher forcing training. This supervisor network is responsible for learning the inherent temporal dependencies in time series data. By training the supervisor jointly with the autoencoder, we aim to fulfill both objectives of time series generation. This integration is crucial because time series data exhibit temporal characteristics such as trends, seasonality, noise, and periodicity. Without capturing these essential features, the synthetic samples produced would be inadequate and of limited practical value.

\subsection{Autoregressive Refinement.}

By integrating AAEs with autoregressive learning, we can not only estimate the true distribution \( p(\mathbf{X}_{1:T}) \) using a learned probability density function \( \hat{p}(\mathbf{X}_{1:T}) \), but also capture the conditional density function \( \hat{p}(X_t|\mathbf{X}_{1:t-1}) \), which approximates the true \( p(X_t|\mathbf{X}_{1:t-1}) \) at any specific time step \( t \). AVATAR achieves this by training a supervisor network in parallel with the autoencoder, thereby fulfilling both objectives. The autoencoder and supervisor are trained jointly with a combined loss function \( \mathcal{L}_{AE} \) that includes \( \mathcal{L}_{R_{\text{joint}}} \), \( \mathcal{L}_{\text{Ad}} \), \( \mathcal{L}_{S} \), and \( \mathcal{L}_{D} \):

\begin{equation}
\label{eq:one}
\mathcal{L}_{AE} = \mathcal{L}_{R_{\text{joint}}} + \mathcal{L}_{\text{Ad}} + \mathcal{L}_{S} + \mathcal{L}_{D}
\end{equation}

In Equation \eqref{eq:one}, the supervised loss \( \mathcal{L}_{S} \) represents a novel contribution used to train the supervisor network. This loss is designed to train the autoencoder-supervisor as a unified network to predict time step \( t \) using inputs from time steps 1 to \( t-1 \) and also from time steps 1 to \( t-2 \). This dual approach enables the integrated autoencoder-supervisor network to learn temporal dependencies without overfitting to immediate next-step dynamics. Specifically, in closed-loop mode, the supervisor receives sequences of the autoencoder's output \( {X}^{AE}_{1:t-1} \) and \( {X}^{AE}_{1:t-2} \) to predict the vector at time step \( t \), denoted as \( {X}^{AE}_t \). Gradients are calculated from a loss that measures the divergence between the distributions \( p({X}^{AE}_t|{X}^{AE}_{1:t-1}) \) and \( \hat{p}({X}^{AE}_t|{X}^{AE}_{1:t-1}) \), as well as between \( p({X}^{AE}_t|{X}^{AE}_{1:t-2}) \) and \( \hat{p}({X}^{AE}_t|{X}^{AE}_{1:t-2}) \). Using maximum likelihood estimation, this leads to the following supervised loss:

\begin{align}
\mathcal{L}_S = & \ \mathbb{E}_{x_{1:T} \sim p} \left[ \sum_t \left\| X^{AE}_t - s(X^{AE}_{t-1}) \right\|_2 \right] \\
& + \mathbb{E}_{x_{1:T} \sim p} \left[ \sum_t \left\| X^{AE}_t - s(X^{AE}_{t-2}) \right\|_2 \right]
\end{align}

where \( s \) is the supervisor function and \( X^{AE} \) is the output of the autoencoder for input data \( X \). Through joint training of the autoencoder and supervisor, the framework learns to capture the temporal dynamics of time-series data.

\subsection{Distribution Loss.}
The adversarial feedback from the discriminator to the encoder is insufficient for guiding the encoder to effectively learn the prior Gaussian distribution. To address this limitation, we propose a novel distribution loss \( \mathcal{L}_{D} \), which facilitates the encoder's ability to better approximate the target probability distribution by explicitly capturing the differences in the mean and standard deviation (std) between batches of latent code \( Z \) and the prior distribution \( \hat{Z} \):

\begin{equation}
\mathcal{L}_D = \mathcal{L}_{Mean} + \mathcal{L}_{Std}
\end{equation}

Although various statistical measures can be used to describe a distribution, in the context of our framework, metrics such as minimum, maximum, median, mode, and quartiles are not applicable. This is due to the fact that the AVATAR framework automatically normalizes the data within the range of 0 to 1, making the minimum and maximum values redundant as all data points fall within this standardized interval. Additionally, because we employ a Gaussian distribution as the prior, the median and mode align with the mean. Quartiles, meanwhile, introduce unnecessary noise and do not provide significant insights into the distribution’s characteristics. Consequently, the only metrics of relevance in this context are the mean and std, as these effectively capture the properties of the Gaussian distribution.

\begin{equation}
\mathcal{L}_{\text{Mean}} = \mathbb{E}_{z_{1:T} \sim q, \hat{z}_{1:T} \sim \hat{q}} \left[ \sum_t \left| \frac{1}{N} \sum_{n=1}^N \mathbf{Z}_{t_n} - \frac{1}{N} \sum_{n=1}^N \mathbf{\hat{Z}}_{t_n} \right| \right]
\end{equation}

In this formulation, $\mathcal{L}_{\text{Mean}}$ computes the mean absolute error (MAE) between the means of a batch of latent code $Z$ and prior distribution data $\hat{Z}$. We consider sequences of temporal data, denoted as $\mathbf{Z}_{1:T}$ or $\mathbf{\hat{Z}}_{1:T}$, drawn from a joint distribution $q$ or $\hat{q}$, where each sample is indexed by $n \in \{1, \ldots, N\}$, and the batch is represented as $\mathcal{B} = \{\mathbf{Z}_{n,1:T_n}\}_{n=1}^{N}$.

\begin{align}
\mathcal{L}_{\text{Std}} = & \, \mathbb{E}_{z_{1:T} \sim q, \hat{z}_{1:T} \sim \hat{q}} \left[ \sum_t \left| \frac{1}{N} \sum_{n=1}^N ( \mathbf{Z}_{t_n} - \overline{\mathbf{Z_t}})^2 \right. \right. \nonumber \\
& \left. \left. - \frac{1}{N} \sum_{n=1}^N(\mathbf{\hat{Z}}_{t_n} - \overline{\mathbf{\hat{Z_t}}})^2 \right| \right]
\end{align}

Similarly, $\mathcal{L}_{\text{Std}}$ measures the MAE between the std of a batch of $Z$ and $\hat{Z}$. In this context, $\overline{\mathbf{Z}}$ and $\overline{\mathbf{\hat{Z}}}$ represent the means of ${Z}$ and $\hat{{Z}}$, respectively, for a batch of data.

\subsection{Regularized GRU.}
The architecture of each network in the AVATAR framework significantly impacts performance. Based on previous research in time series generation \cite{seriesgan}, we found that GRU outperforms LSTM and RNN, and thus we chose GRU layers for all four networks: the encoder, decoder, supervisor, and discriminator. GRUs consist of two gates: a reset gate and an update gate, which regulate the flow of information and enable the network to decide what information to retain or discard. This simplified structure allows GRUs to efficiently manage sequential data while maintaining strong predictive power, especially in cases where memory of prior inputs is critical. However, GRU layers can be further enhanced by incorporating additional components. To improve performance, we combine batch normalization with GRU layers. As a result, each layer consists of both a batch normalization and a GRU unit, with multiple such layers used for the encoder, decoder, and supervisor. For the discriminator, we avoid using batch normalization, as it would make the discriminator overly powerful, disrupting the balance between the discriminator and encoder, leading to the discriminator’s dominance. To further maintain this balance, we train the encoder with twice as many iterations as the discriminator to prevent the latter from overpowering the former.

\subsection{Joint Training.}
The training process of the AVATAR framework consists of three distinct stages. In the first stage, the autoencoder is trained independently, relying solely on a reconstruction loss $\mathcal{L}_{R}$. At this point, neither the supervisor nor the discriminator is present, and the autoencoder functions as a conventional autoencoder:
\begin{equation}
\mathcal{L}_{R} = \mathbb{E}_{x_{1:T}\sim p} \left[ \sum_t \|\mathbf{X}_t - \mathbf{{X}^{AE}}_t\|_2 \right] 
\end{equation}

where $X$ denotes the input data and ${X}^{AE}$ represents the autoencoder’s output.

In the second stage, the supervisor is trained using the previously defined $\mathcal{L}_{S}$ loss.

The third stage consists of two distinct phases. In the initial phase, the integrated autoencoder and supervisor are trained jointly, guided by the loss function \( \mathcal{L}_{AE} \). This loss function is composed of four sub-components as discussed earlier, two of which are: (i) \( \mathcal{L}_{R_{joint}} \), representing the joint reconstruction loss for the integrated network, and (ii) \( \mathcal{L}_{Ad} \), which captures the conventional adversarial feedback loss.

\begin{align}
\mathcal{L}_{R_{joint}} = & \mathbb{E}_{x_{1:T}\sim p} \left[ \sum_t \|\mathbf{X}_t - \mathbf{{X}^{AE}}_t\|_2 \right] 
\nonumber \\
& + \mathbb{E}_{x_{1:T}\sim p} \left[ \sum_t \|\mathbf{X}_t - \mathbf{s({X}^{AE}}_t)\|_2 \right] 
\end{align}

In the second phase of the third stage, the discriminator is trained to distinguish between samples from the prior distribution and the aggregated posterior latent representation. This joint training, through the combined loss function $\mathcal{L}_{AE}$, effectively fulfills the dual objectives of generating time series data and achieving enhanced stability in the learning process. From an optimization perspective, no hyperparameters for loss weighting have been applied to any of the components of the combined loss functions, including \( \mathcal{L}_{AE}, \mathcal{L}_{R_{joint}}, \mathcal{L}_S \), and \( \mathcal{L}_D \). As a result, all components contribute equally to the optimization process.

\section{Experiments and Results.}
The Python implementation of the AVATAR framework, along with a tutorial Jupyter notebook demonstrating its usage and the results of experiments, is available online. This implementation incorporates all the innovations discussed, as well as the hyperparameters necessary to achieve optimal results\footnote{The AVATAR framework is accessible here: \href{https://github.com/samresume/AVATAR}{https://github.com/samresume/AVATAR}}.

\subsection*{Baseline Methods.}

The AVATAR framework has been evaluated against several state-of-the-art data augmentation techniques, including TimeGAN \cite{timegan}, Standard GAN \cite{gan}, Standard AAE \cite{aae}, Teacher Forcing (T-Forcing) \cite{Tforcing}, and Professor Forcing (P-Forcing) \cite{Pforcing}, covering both GAN-based and autoregressive-based approaches. To ensure a fair and rigorous evaluation, identical hyperparameters are applied across all methods. These include the number of iterations, the type of recurrent neural network, the number of layers, batch size, and hidden dimensions.

\subsection*{Benchmark Datasets.}

We evaluate AVATAR’s effectiveness using time series datasets that showcase a wide range of features, such as periodicity, noise intensity, and trend patterns. These datasets are selected based on various combinations of these features.

\begin{itemize}

    \item \textbf{Energy:} The UCI Appliances Energy Prediction dataset \cite{UCI_Appliances_Energy} is distinguished by irregular periodic patterns, noise, a high number of dimensions (28), and correlated features, presenting significant challenges for modeling and analysis. It comprises multivariate, continuous-valued measurements that capture various temporal attributes, all recorded at closely spaced intervals.
    
    \item \textbf{Google Stock:} Stock price data is continuous but lacks periodicity, with correlated features. We utilize historical daily stock data from Google between 2004 and 2019, including variables such as trading volume, daily highs and lows, opening prices, closing prices, and adjusted closing values.
    
    \item \textbf{Sinusoidal Data:} We generate multivariate sinusoidal time series, where each series has distinct frequencies ($\eta$) and phases ($\theta$), resulting in continuous, periodic signals with independent features. For each dimension $i$ (ranging from 1 to 4), the function is represented as $x_i(t) = \sin(2\pi\eta t + \theta)$, where $\eta$ is drawn from a uniform distribution $U[0, 1]$ and $\theta$ from $U[-\pi, \pi]$.
\end{itemize}

For the Energy and Stocks series, we slice each continuous time series into samples using a slicing window of size 24, moving the window forward by one time step to extract each subsequent sample throughout the series.

\subsection*{Evaluation Criteria.}

Evaluating the performance of GANs and AAEs comes with significant challenges. Methods based on likelihood, such as Parzen window estimates \cite{ganevaluation1, ganevaluation2}, can produce misleading outcomes. While human evaluation is typically viewed as the most reliable way to assess quality, it is both impractical and expensive. As a result, this work evaluates performance based on three recent evaluation metrics, incorporating both qualitative and quantitative assessments of the generated data. These include resemblance score, predictive fidelity score, and distributional alignment analysis using t-SNE and PCA visualization.

\subsection{Resemblance Score.} To quantitatively assess similarity, we train a time series classifier based on LSTM to distinguish between sequences from the original and synthetic datasets. The LSTM classifier is trained to separate these two categories in a typical supervised learning setup. The classification error on a separate test set is used to derive a quantitative measure, which is then subtracted from 0.5, setting the optimal score to 0 rather than 0.5. This metric emphasizes the utility of the generated data in downstream classification tasks. To ensure reliable and robust results, each experiment is repeated 10 times. The mean and std of these observations are reported in Table \ref{tbl:scores} for a comprehensive comparison. Based on the results presented in Table \ref{tbl:scores}, AVATAR significantly outperforms the baseline techniques, including the state-of-the-art TimeGAN, across the specified three datasets. Specifically, AVATAR reduces the resemblance score by 46.86\% compared to TimeGAN. Both GANs and AAEs fail to capture the time series characteristics effectively, leading to the production of unrealistic data. This highlights the importance of incorporating innovative techniques to enhance their performance in time series data generation. Additionally, T-forcing achieved a reliable score, underscoring the effectiveness of autoregressive learning for time series generation.
\begin{table}
\centering
\caption{Resemblance scores and Predictive Fidelity scores for various time series generation techniques across different datasets.}
\label{tbl:scores}
\scriptsize
\begin{tabular}{cccc}
\hline
\multicolumn{4}{c}{\textit{Resemblance Score}} \\ \hline
\textbf{} & \textbf{Stocks} & \textbf{Sines} & \textbf{Energy}\\ \hline

\textbf{AVATAR} & \textbf{0.129 ± 0.005} & \textbf{0.064 ± 0.033} & \textbf{0.298 ± 0.002} \\ 

TimeGAN & 0.238 ± 0.007 & 0.166 ± 0.048 & 0.447 ± 0.002 \\ 

GAN & 0.489 ± 0.001 & 0.499 ±  0.001 & 0.5 ± 0.0 \\ 

AAE & 0.465 ± 0.011 & 0.487 ± 0.009 & 0.499 ± 0.001 \\ 

T-Forcing & 0.174 ± 0.064 & 0.152 ± 0.140 & 0.411 ± 0.063 \\ 

P-Forcing & 0.5 ± 0.0 & 0.441 ± 0.007 & 5.0 ± 0.0 \\ \hline

\multicolumn{4}{c}{\textit{Predictive Fidelity Score}} \\ \hline

\textbf{} & \textbf{Stocks} & \textbf{Sines} & \textbf{Energy} \\ \hline

\textbf{AVATAR} & \textbf{0.043 ± 0.001} & \textbf{0.095 ± 0.001} & \textbf{0.158 ± 0.003} \\ 

TimeGAN & 0.049 ± 0.001 & 0.128 ± 0.003 & 0.206 ± 0.001 \\ 

GAN & 0.100 ± 0.016 & 0.139 ± 0.006 & 0.257 ± 0.003 \\ 

AAE & 0.053 ± 0.006 & 0.121 ± 0.003 & 0.302 ± 0.004 \\ 

T-Forcing & 0.047 ± 0.002 & 0.112 ± 0.002 & 0.188 ± 0.001 \\ 

P-Forcing & 0.070 ± 0.002 & 0.115 ± 0.001 & 0.281 ± 0.006 \\ \hline

\end{tabular}
\end{table}

\subsection{Predictive Fidelity Score.} We expect AVATAR to successfully capture the conditional distributions over time. To measure this, we train an LSTM model on the synthetic dataset for sequence prediction, focusing on forecasting the next-step temporal vectors for each input sequence. The performance of the model is then evaluated on the original dataset, with accuracy assessed using MAE. This metric demonstrates the quality of the synthetic data in predictive tasks, which is a key objective in time series analysis. Similar to the evaluation of the resemblance score, we repeated the experiments 10 times to ensure a robust assessment. As shown in Table \ref{tbl:scores}, AVATAR demonstrates superior results in terms of the predictive capability of synthetic data compared to the baseline techniques. Specifically, AVATAR reduces the predictive error by 20.44\% in comparison to TimeGAN. Furthermore, AVATAR exhibits significantly greater stability compared to TimeGAN and GAN-based techniques, consistently yielding the same results across different training sessions.  Additionally, T-forcing achieved a stronger performance relative to the other baseline models, highlighting the effectiveness of autoregressive training for synthetic data generation.

\begin{figure}
\centering
\subfloat{\includegraphics[width=0.163\textwidth]{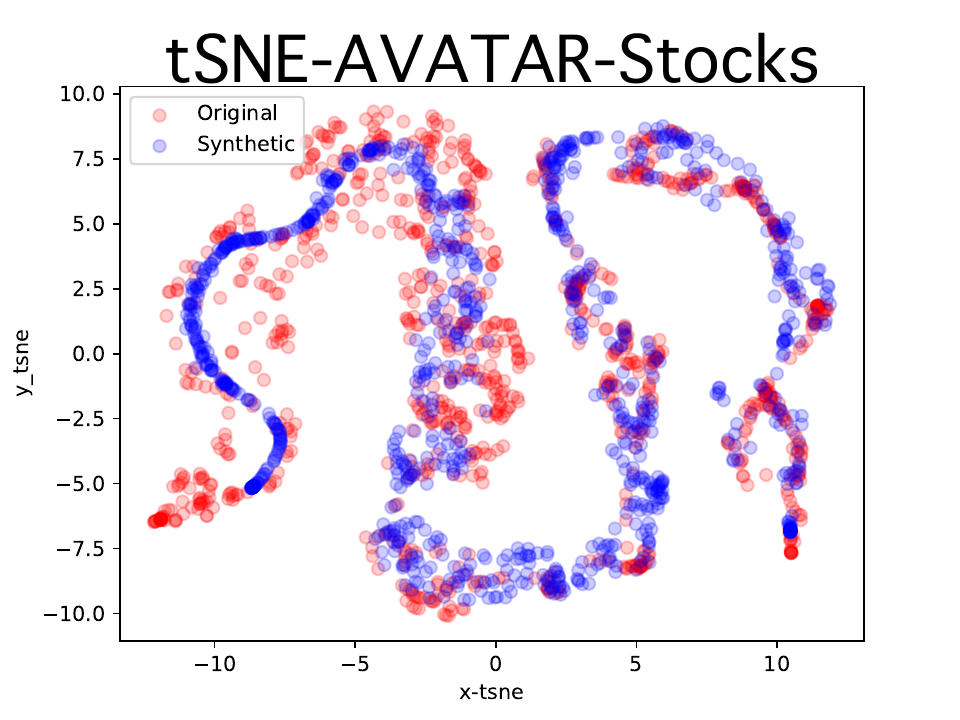}}\hfill
\subfloat{\includegraphics[width=0.163\textwidth]{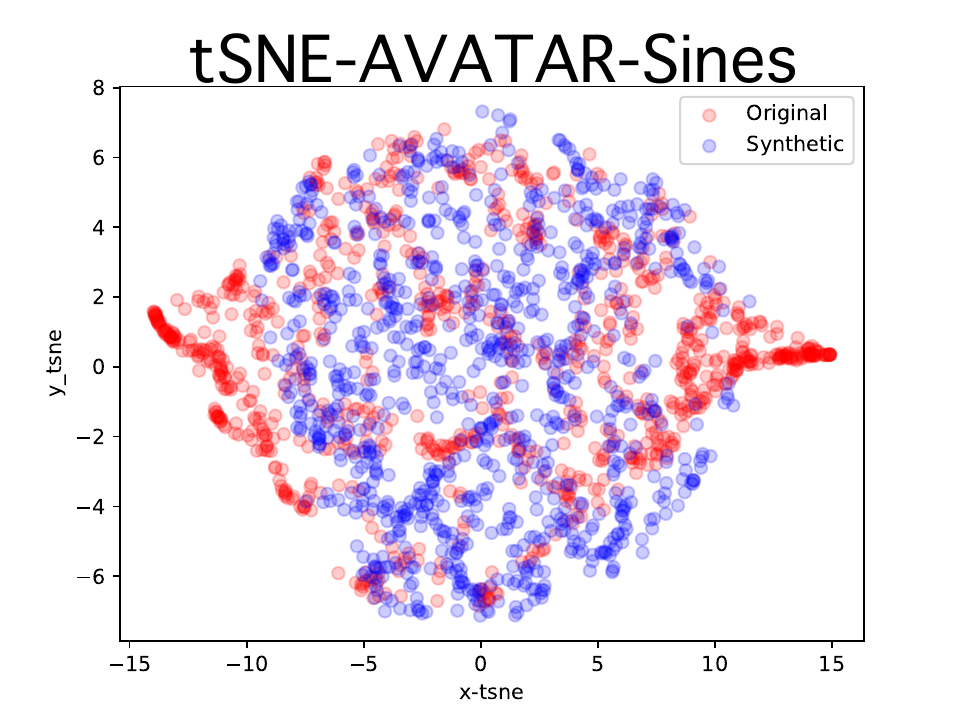}}\hfill
\subfloat{\includegraphics[width=0.163\textwidth]{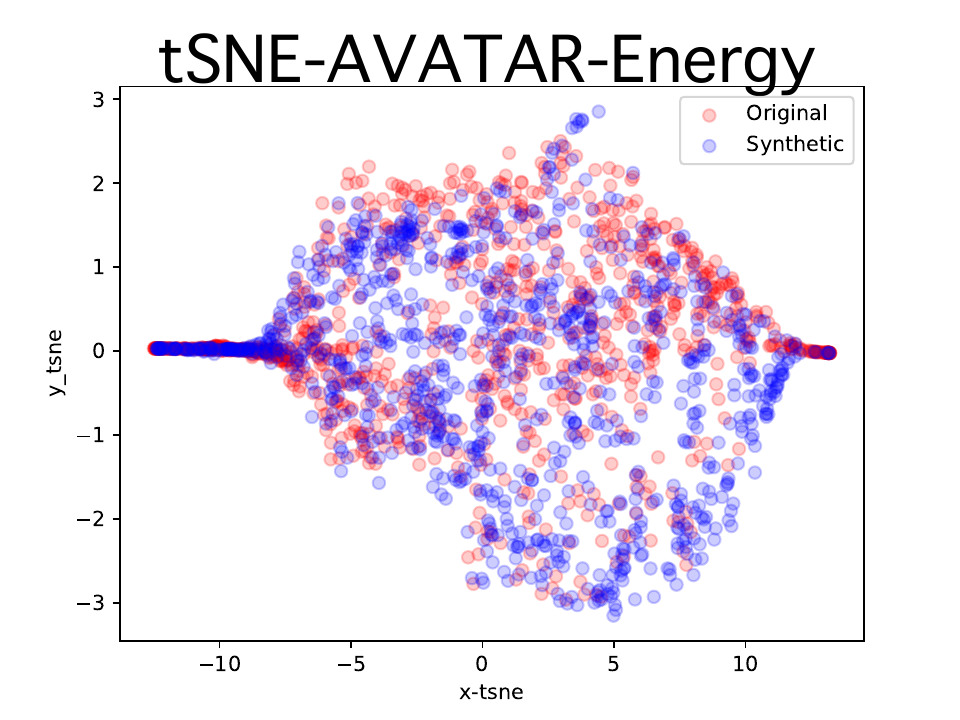}}\hfill

\vspace{-1em}

\subfloat{\includegraphics[width=0.163\textwidth]{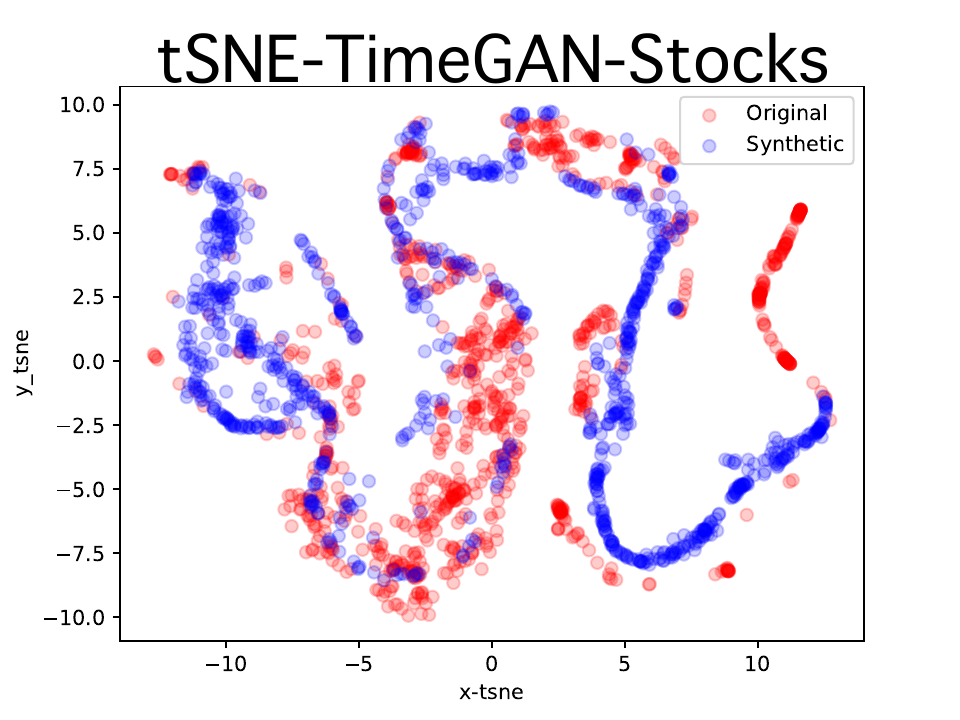}}\hfill
\subfloat{\includegraphics[width=0.163\textwidth]{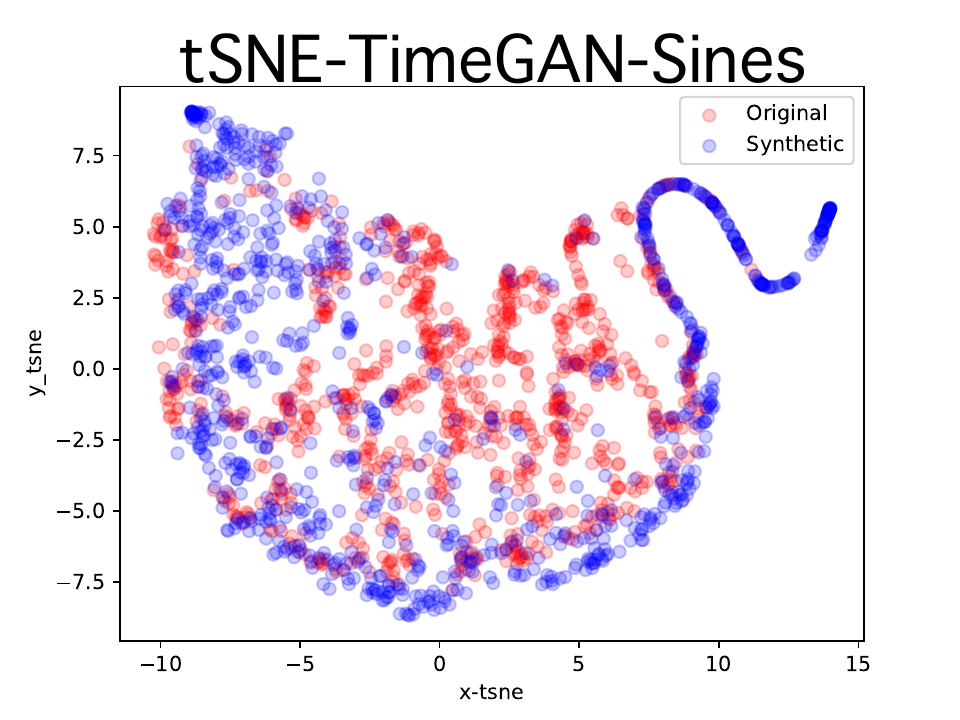}}\hfill
\subfloat{\includegraphics[width=0.163\textwidth]{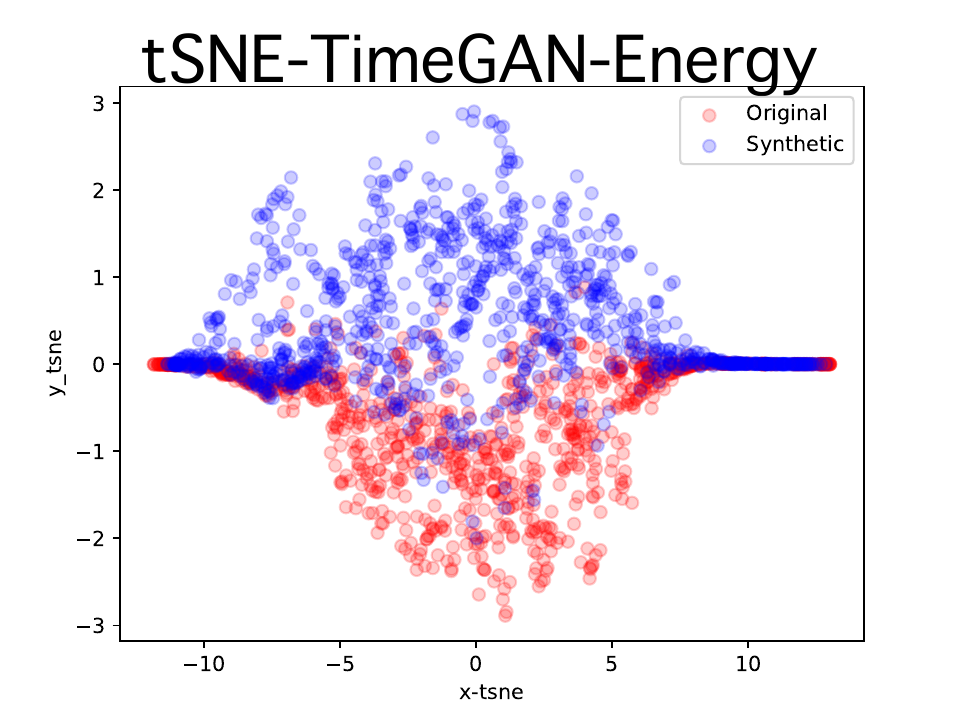}}\hfill

\caption{t-SNE visualizations illustrate the distribution alignment between original (red) and synthetic (blue) data samples generated by AVATAR and TimeGAN across three datasets.}
\label{fig:tsne}
\end{figure}

\subsection{Distributional Alignment.} 
\begin{figure}
\centering
\subfloat{\includegraphics[width=0.163\textwidth]{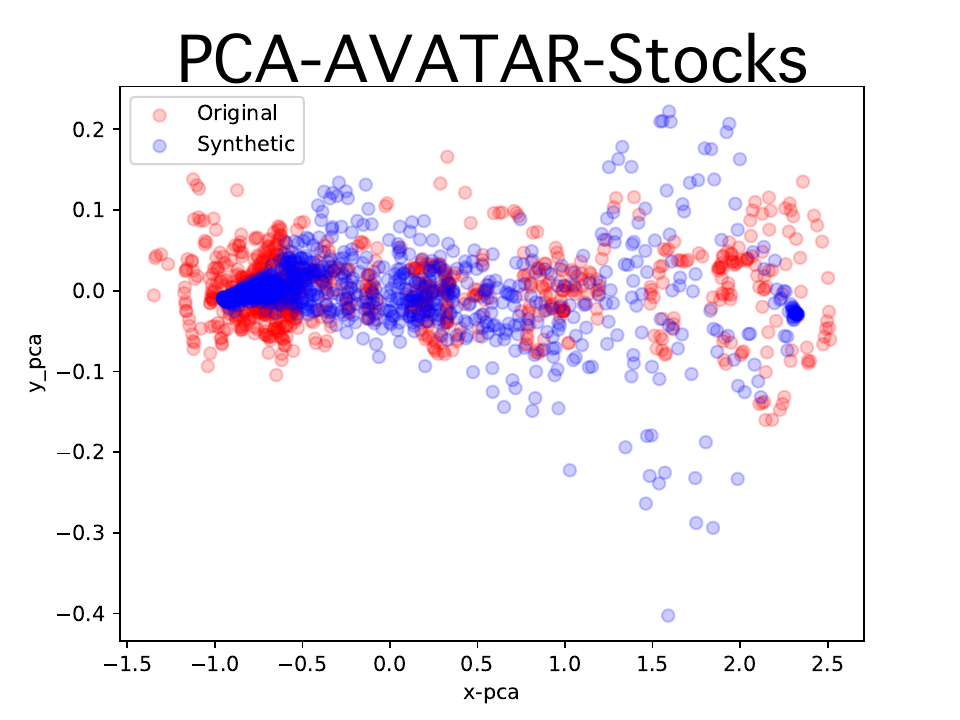}}\hfill
\subfloat{\includegraphics[width=0.163\textwidth]{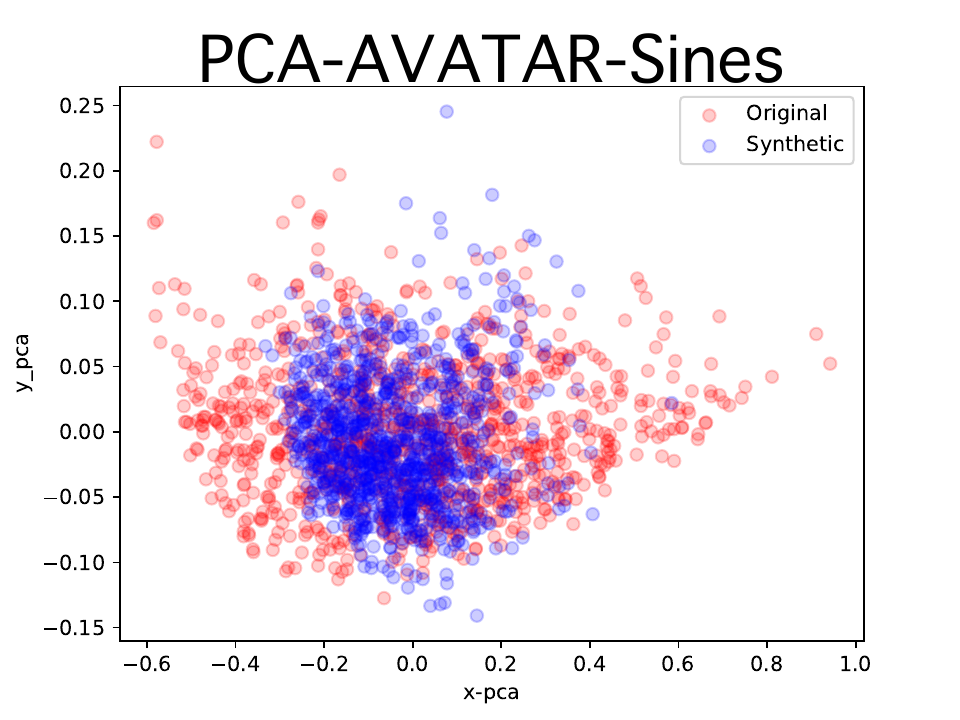}}\hfill
\subfloat{\includegraphics[width=0.163\textwidth]{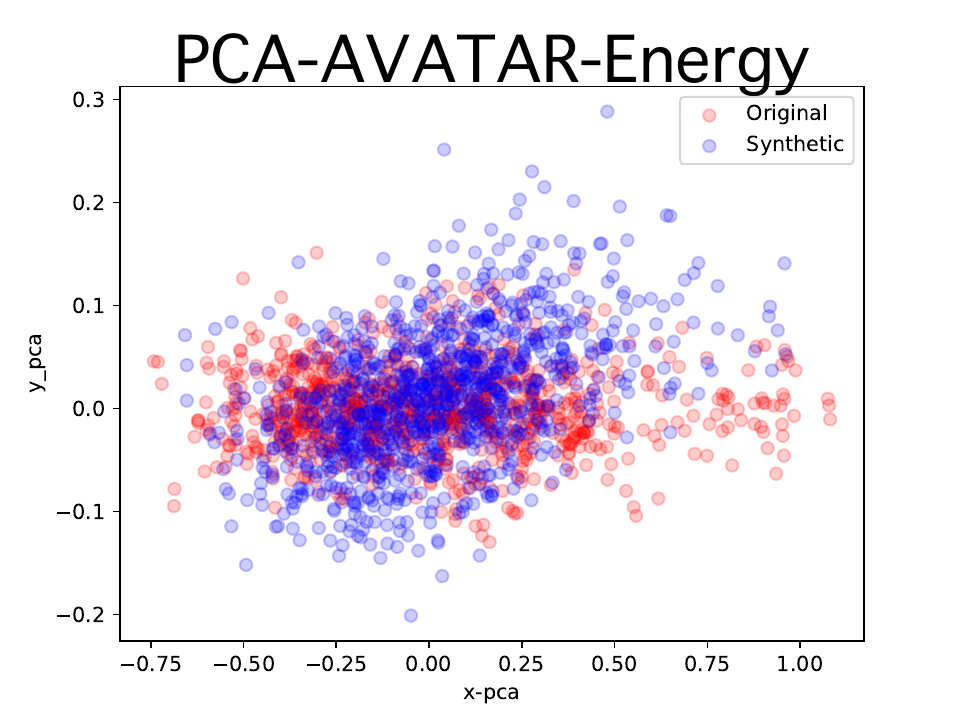}}\hfill

\vspace{-1em}

\subfloat{\includegraphics[width=0.163\textwidth]{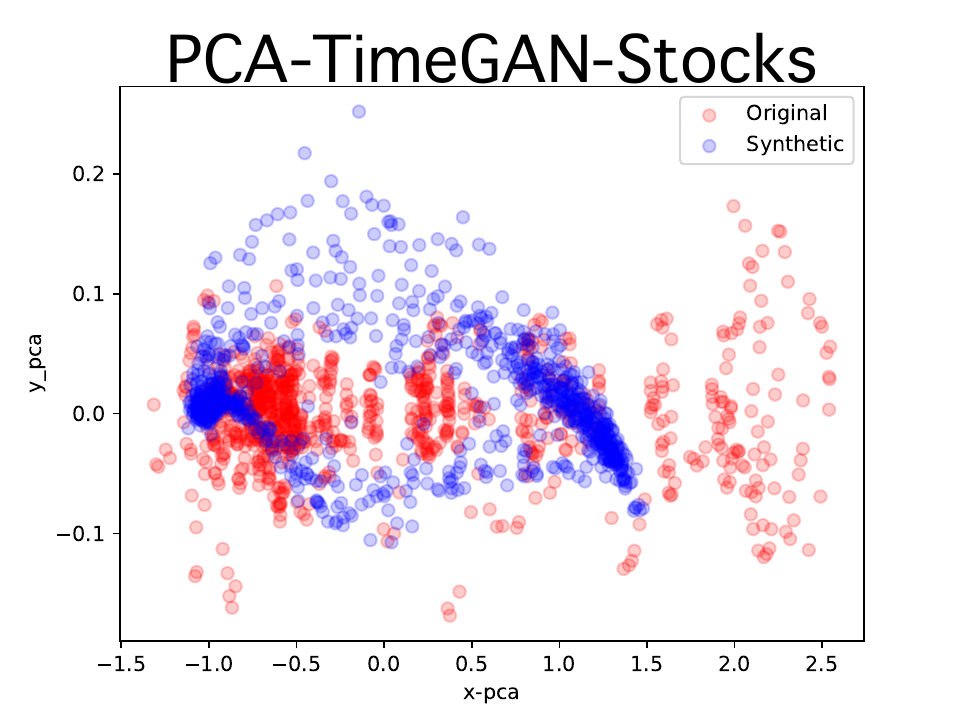}}\hfill
\subfloat{\includegraphics[width=0.163\textwidth]{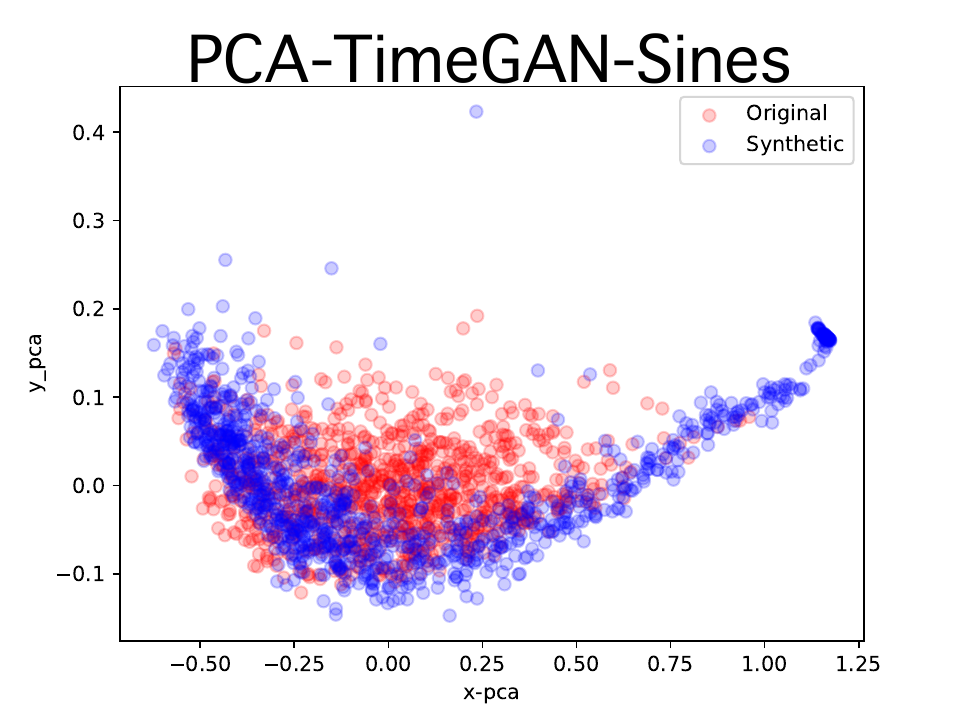}}\hfill
\subfloat{\includegraphics[width=0.163\textwidth]{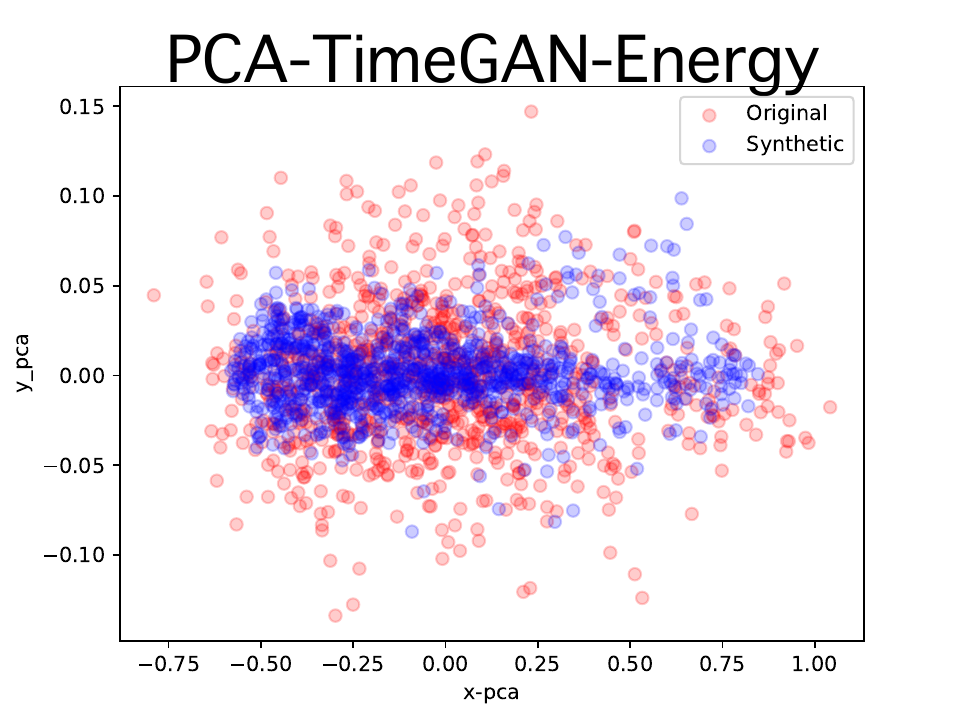}}\hfill

\caption{PCA visualizations depict the alignment in distribution between the original (red) and synthetic (blue) data samples produced by AVATAR and TimeGAN across three datasets.}
\label{fig:pca}
\end{figure}

We utilize t-SNE \cite{tsne} and PCA \cite{pca} analyses on both original and synthetic datasets by reducing the temporal dimension for visualization. By doing so, we can qualitatively evaluate how well the distribution of the generated samples aligns with that of the original data in a two-dimensional space. This method provides insight into one of the core goals of a generative model, which is to approximate the probability density function, $\hat{p}(\mathbf{X}{1:T})$, to the true distribution $p(\mathbf{X}{1:T})$. Based on Figures \ref{fig:tsne} and \ref{fig:pca}, the distributional alignment between the original and synthetic data generated by AVATAR is exceptional across all three datasets. This highlights the model’s strength in learning the entire data distribution of the original datasets. Moreover, AVATAR demonstrates a more accurate alignment compared to TimeGAN. This improvement can be attributed to two key factors: first, the use of autoregressive learning, which enhances the overall quality of the synthetic data, and second, the introduction of a novel distribution loss, which enables the aggregated posterior of the latent code to better match the prior Gaussian distribution. Additional PCA and t-SNE plots can be found in Appendix Section \ref{sec:appendixA}.

\subsection{Contribution of Innovations.}

In this section, we analyze the impact of each innovation on improving the performance of the AVATAR framework. To evaluate their contributions, we systematically remove each novelty and rerun the resemblance and predictive fidelity score experiments. As shown in Table \ref{tbl:scores_cont}, removing the joint training mechanism significantly degrades the framework’s performance, as key objectives, including autoregressive learning, rely on joint training through combined loss functions. Similarly, removing autoregressive learning causes a substantial drop in performance, as it addresses the local objective of time series generation by capturing the conditional distribution at each time step. Additionally, excluding the regularized GRU and distribution loss leads to lower scores, as these components facilitate training and ensure alignment between the prior distribution and the latent code.

\begin{table}
\centering
\caption{The contribution of each innovation in enhancing the performance of the AVATAR framework. “w/o” denotes the absence of a specific component, while “AL” refers to Autoregressive Learning, “DL” stands for Distribution Loss, “JT” represents Joint Training, and “RG” stands for Regularized GRU. }
\label{tbl:scores_cont}
\scriptsize
\begin{tabular}{cccc}
\hline
\multicolumn{4}{c}{\textit{Resemblance Score}} \\ \hline
\textbf{} & \textbf{Stocks} & \textbf{Sines} & \textbf{Energy}\\ \hline

\textbf{AVATAR} & \textbf{0.129 ± 0.005} & \textbf{0.064 ± 0.033} & \textbf{0.298 ± 0.001} \\ 

W/o AL & 0.275 ± 0.01 & 0.138 ± 0.04 & 0.401 ± 0.006 \\ 

W/o DL & 0.181 ± 0.003 & 0.097 ± 0.032 & 0.354 ± 0.003 \\ 

W/o JT & 0.334 ± 0.08 & 0.093 ± 0.011 & 0.489 ± 0.001 \\ 

W/o RG & 0.164 ± 0.009 & 0.089 ± 0.035 & 0.330 ± 0.002 \\ \hline

\multicolumn{4}{c}{\textit{Predictive Fidelity Score}} \\ \hline

\textbf{} & \textbf{Stocks} & \textbf{Sines} & \textbf{Energy} \\ \hline

\textbf{AVATAR} & \textbf{0.043 ± 0.001} & \textbf{0.095 ± 0.001} & \textbf{0.158 ± 0.001} \\ 

W/o AL & 0.102 ± 0.001 & 0.198 ± 0.028 & 0.189 ± 0.007 \\ 

W/o DL & 0.069 ± 0.003 & 0.123 ± 0.002 & 0.161 ± 0.001 \\ 

W/o JT & 0.086 ± 0.004 & 0.117 ± 0.001 & 0.204 ± 0.008 \\ 

W/o RG & 0.054 ± 0.001 & 0.145 ± 0.003 & 0.162 ± 0.001 \\ \hline

\end{tabular}
\end{table}

\subsection{Visual Analysis.}

\begin{figure}
\centering
\includegraphics[width=0.50\textwidth]{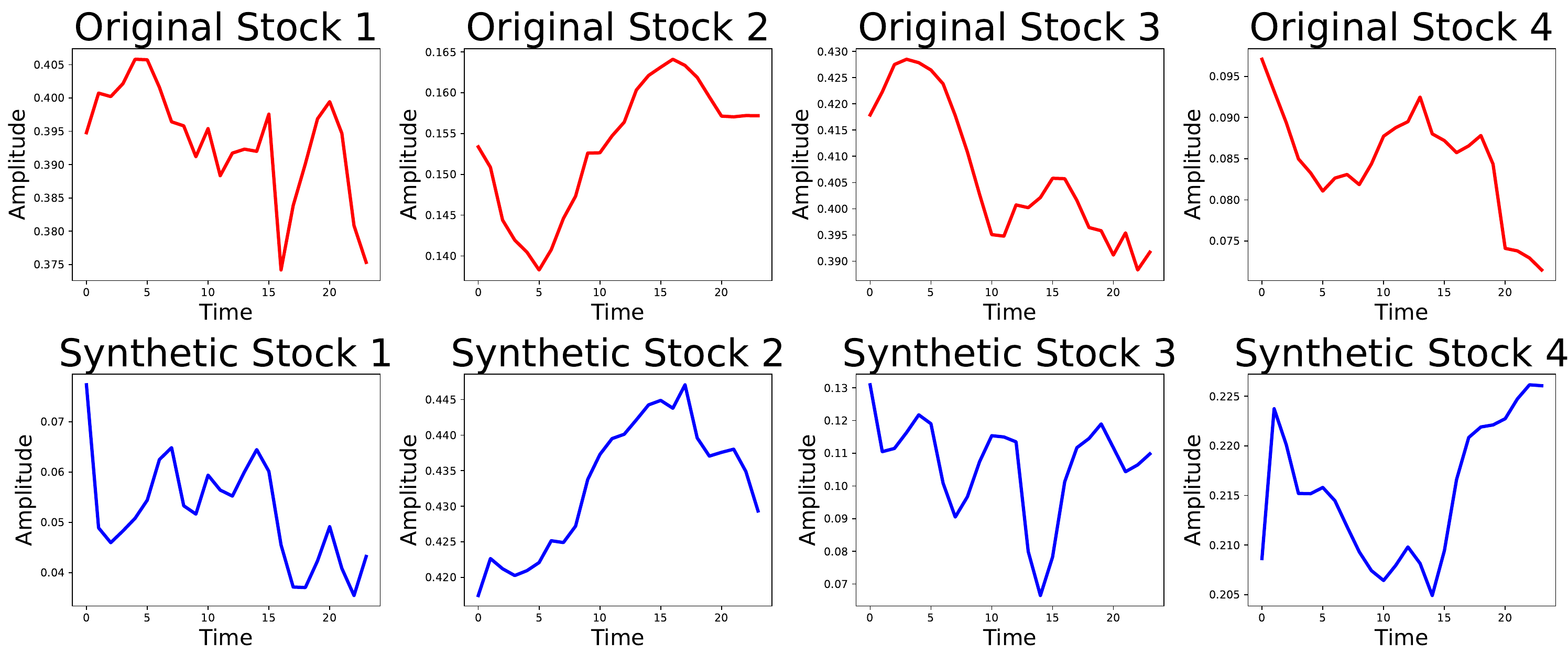}
\caption{The figure presents four randomly selected samples of both original (red) and synthetic data (blue) generated by AVATAR for the Stocks dataset.}
\label{fig:examples}
\end{figure}

As shown in Figure \ref{fig:examples}, which presents four randomly selected original and synthetic data samples generated by AVATAR for the Stocks dataset, the synthetic data demonstrates a strong resemblance to the original data. This high level of similarity plays a key role in the promising results observed across the evaluation metrics. Further visual analyses, including examples of synthetic data for the Sines and Energy datasets, can be found in Appendix Section \ref{sec:appendixB}.

\section{Conclusion.}
In this study, we present AVATAR, an innovative framework built on top of AAEs specifically designed for generating high-quality time series data. AVATAR outperforms TimeGAN and other advanced methods through a comprehensive evaluation using multiple qualitative and quantitative metrics across diverse datasets with varying characteristics. While previous techniques exhibit variability in performance due to instability, leading to inconsistent results with each training session, AVATAR demonstrates significant stability, consistently achieving optimal results after each session. Key innovations in AVATAR include a novel integrated autoregressive training, which effectively captures temporal dynamics, a novel distribution loss that more efficiently models the probabilistic distribution of the data, as well as the incorporation of an improved GRU architecture and joint training. These components collectively drive the superior performance and stability of AVATAR. For future work, we plan to extend the use of AAEs to design a state-of-the-art framework for time series-based missing value imputation.

\section*{Acknowledgments}
This research was funded by the Division of Atmospheric and Geospace Sciences within the Directorate for Geosciences through NSF grants \#2301397, \#2204363, and \#2240022, as well as by the Office of Advanced Cyberinfrastructure within the Directorate for Computer and Information Science and Engineering under NSF grant \#2305781.

\appendix

\section{More PCA and t-SNE Assessment.}

\begin{figure}
\centering
\subfloat{\includegraphics[width=0.163\textwidth]{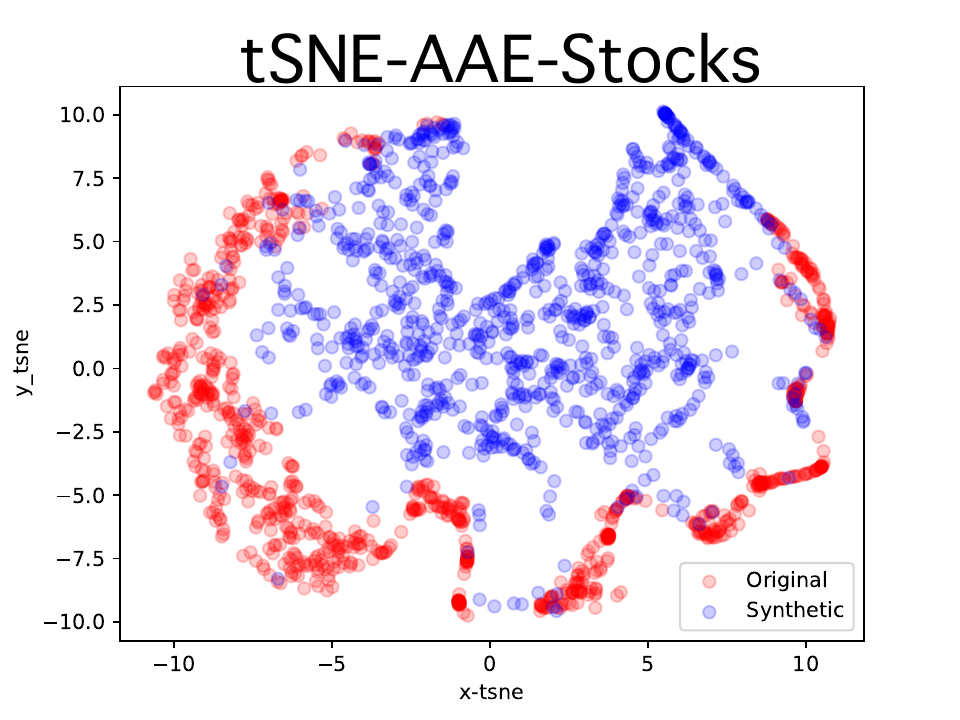}}\hfill
\subfloat{\includegraphics[width=0.163\textwidth]{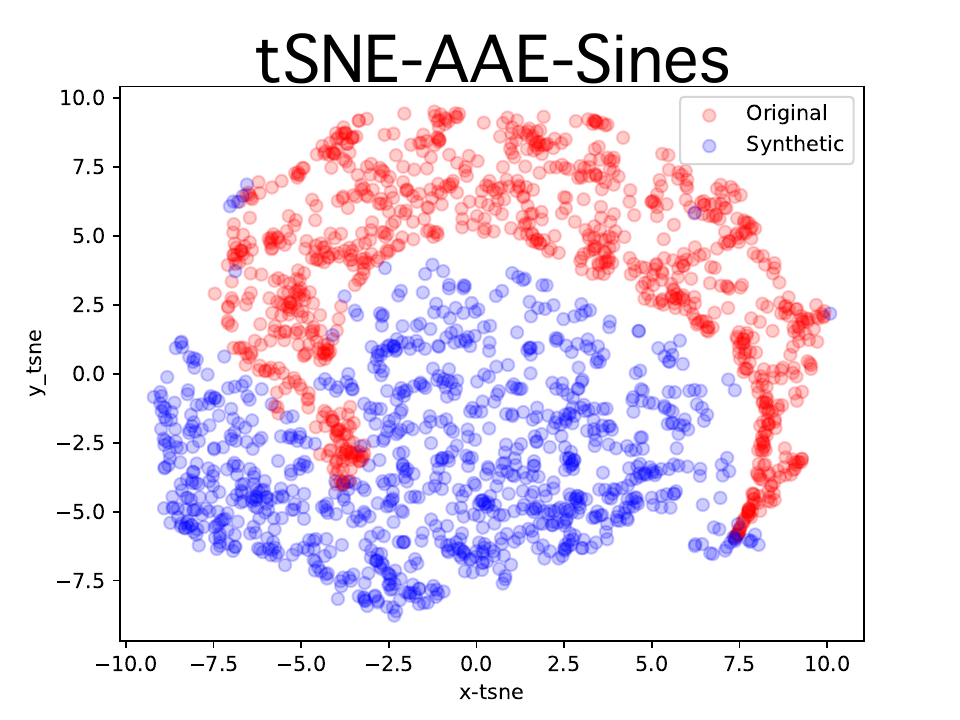}}\hfill
\subfloat{\includegraphics[width=0.163\textwidth]{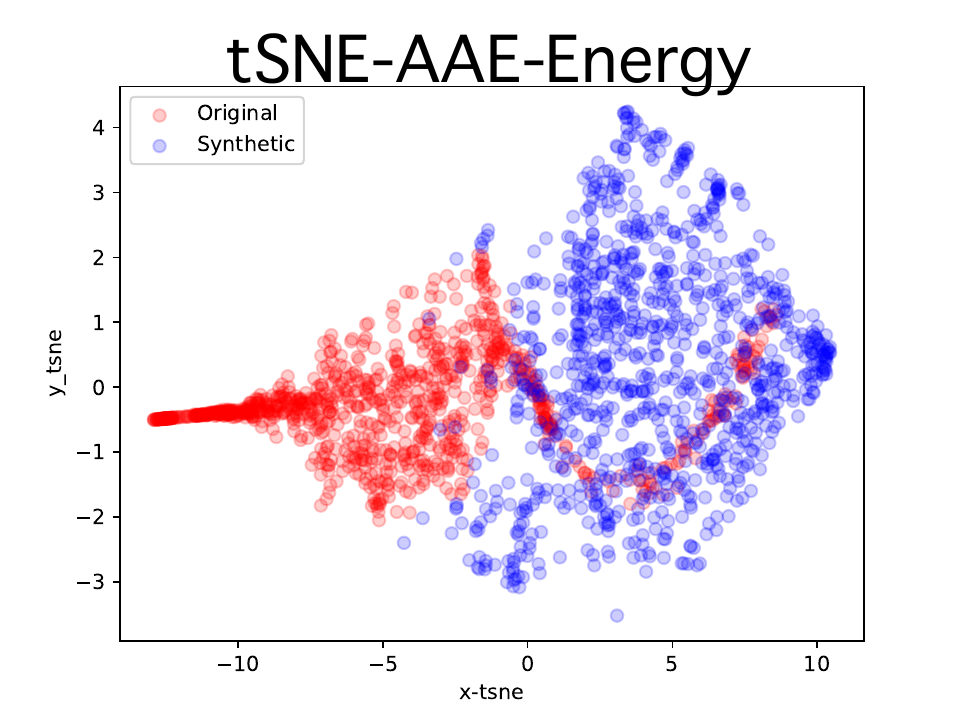}}\hfill

\vspace{-1em}

\subfloat{\includegraphics[width=0.163\textwidth]{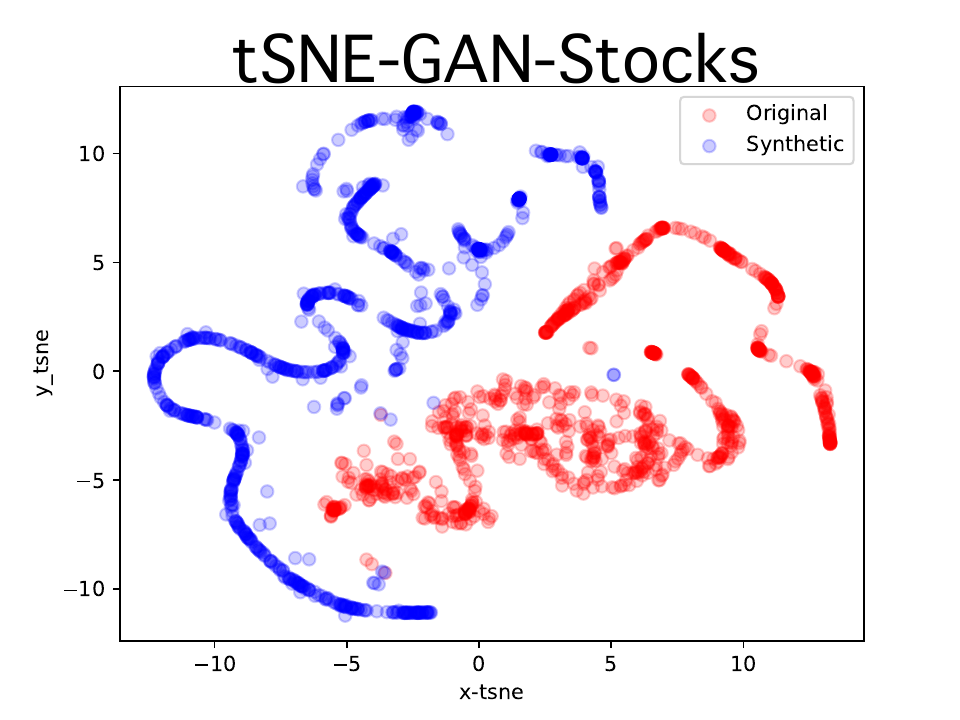}}\hfill
\subfloat{\includegraphics[width=0.163\textwidth]{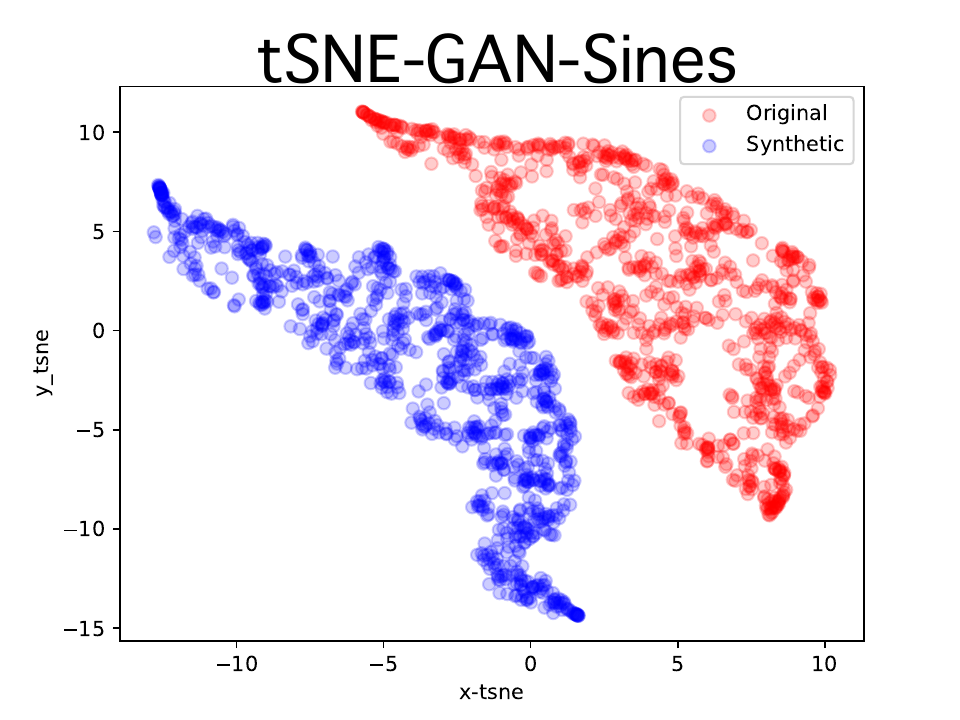}}\hfill
\subfloat{\includegraphics[width=0.163\textwidth]{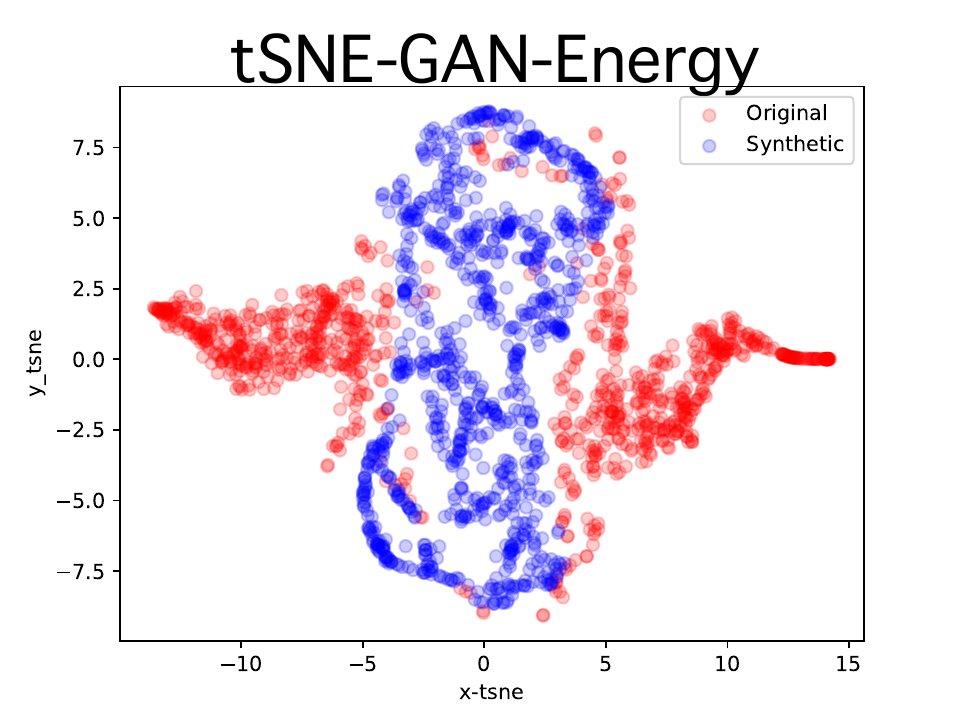}}\hfill

\vspace{-1em}

\subfloat{\includegraphics[width=0.163\textwidth]{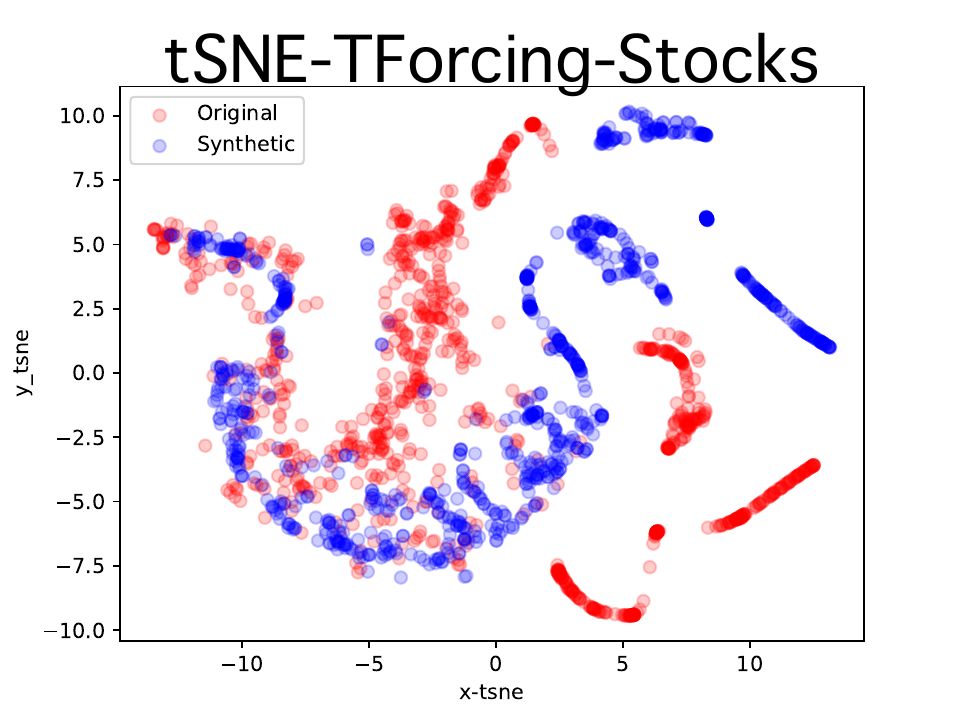}}\hfill
\subfloat{\includegraphics[width=0.163\textwidth]{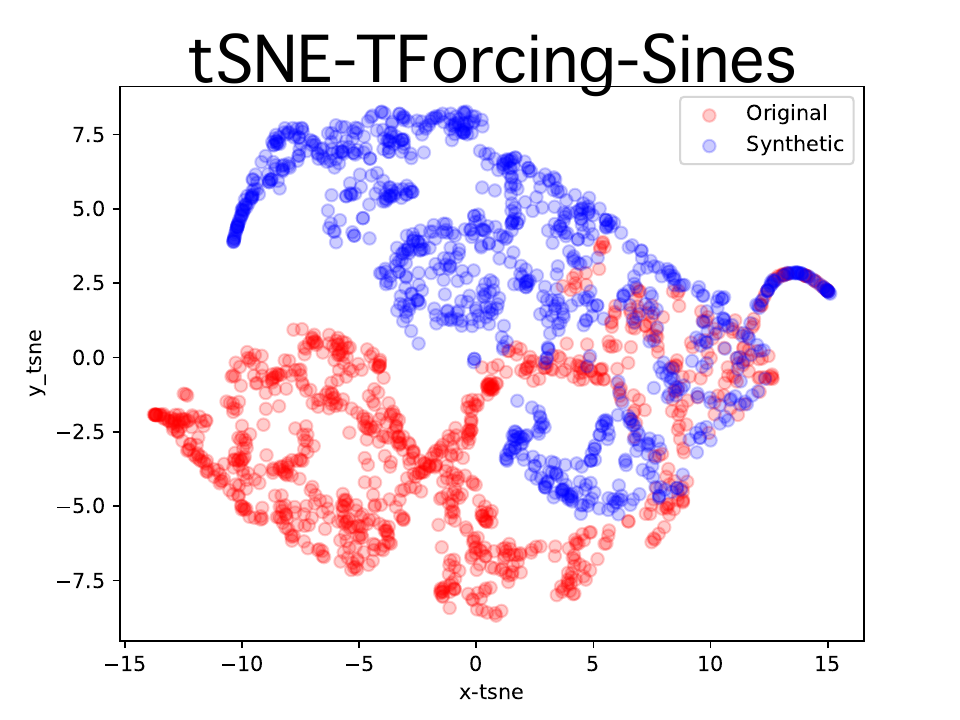}}\hfill
\subfloat{\includegraphics[width=0.163\textwidth]{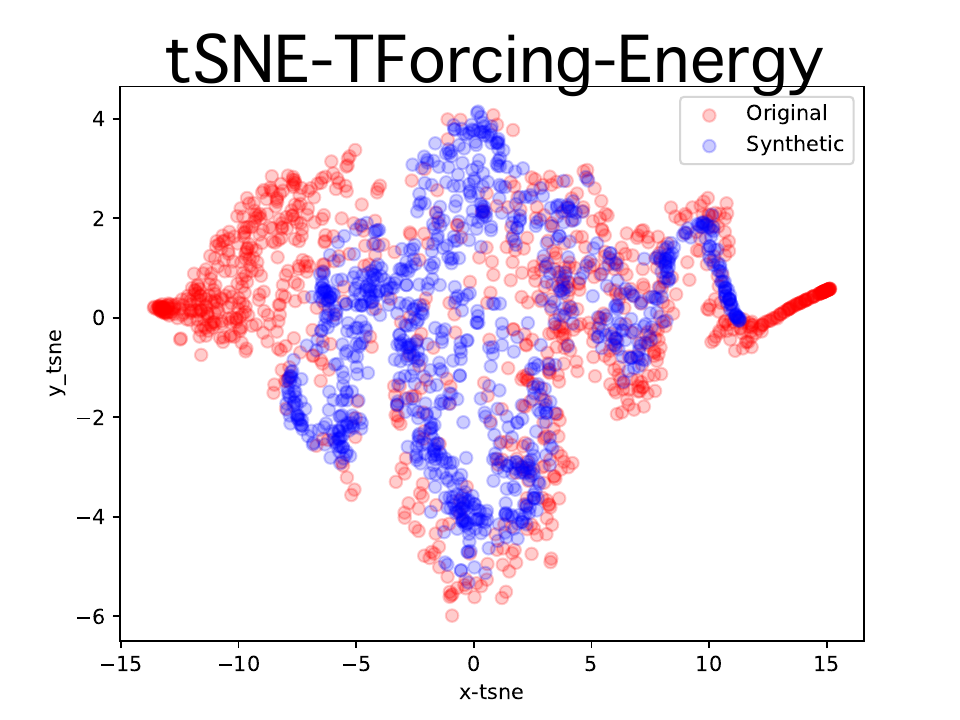}}\hfill

\vspace{-1em}

\subfloat{\includegraphics[width=0.163\textwidth]{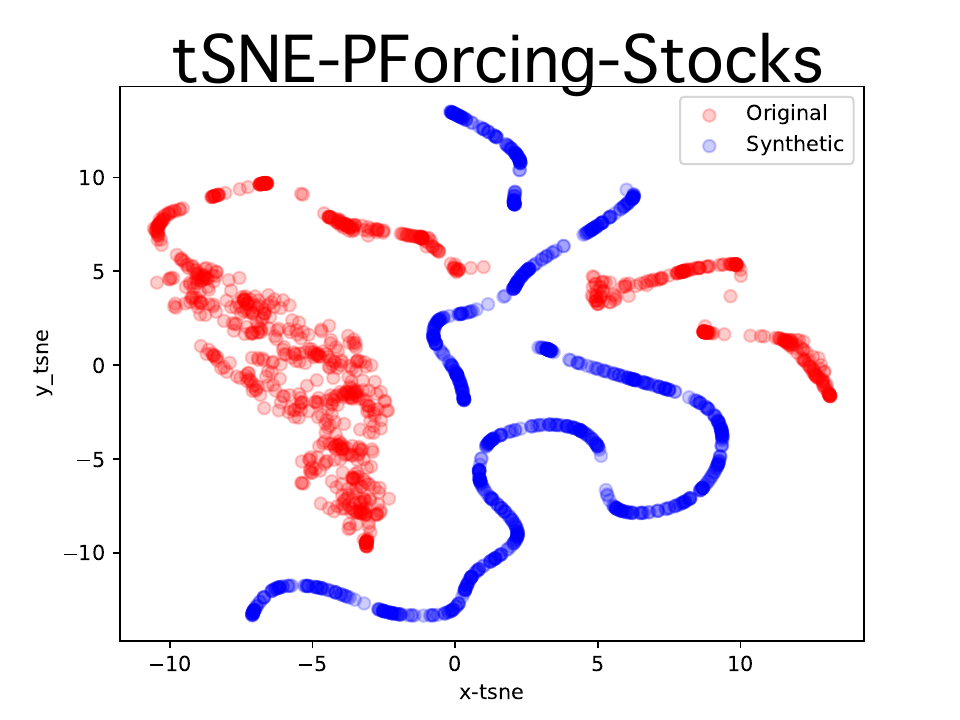}}\hfill
\subfloat{\includegraphics[width=0.163\textwidth]{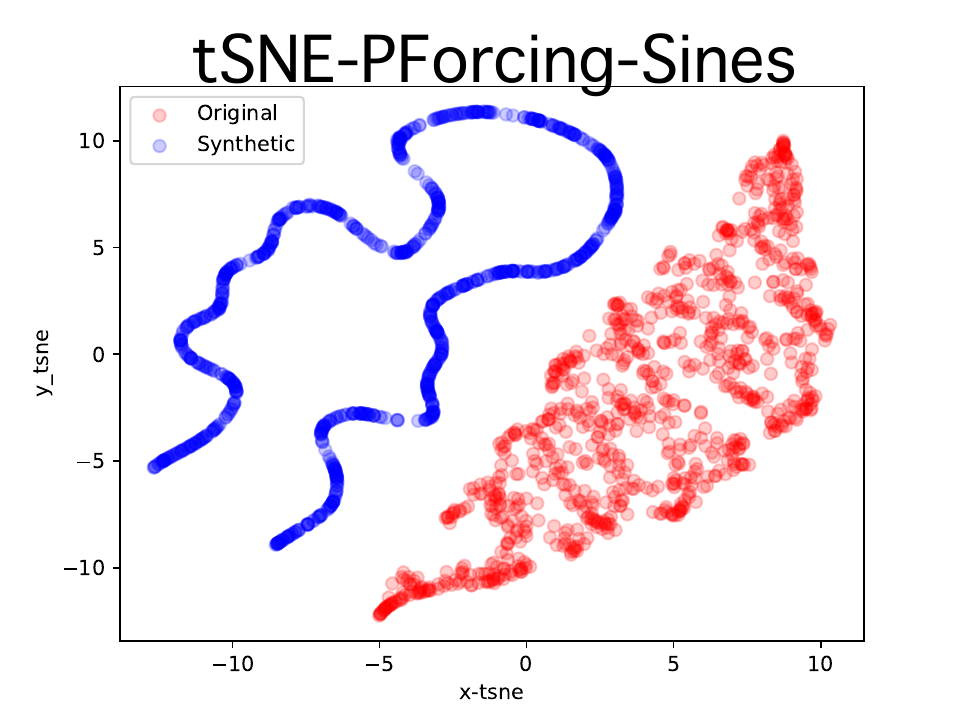}}\hfill
\subfloat{\includegraphics[width=0.163\textwidth]{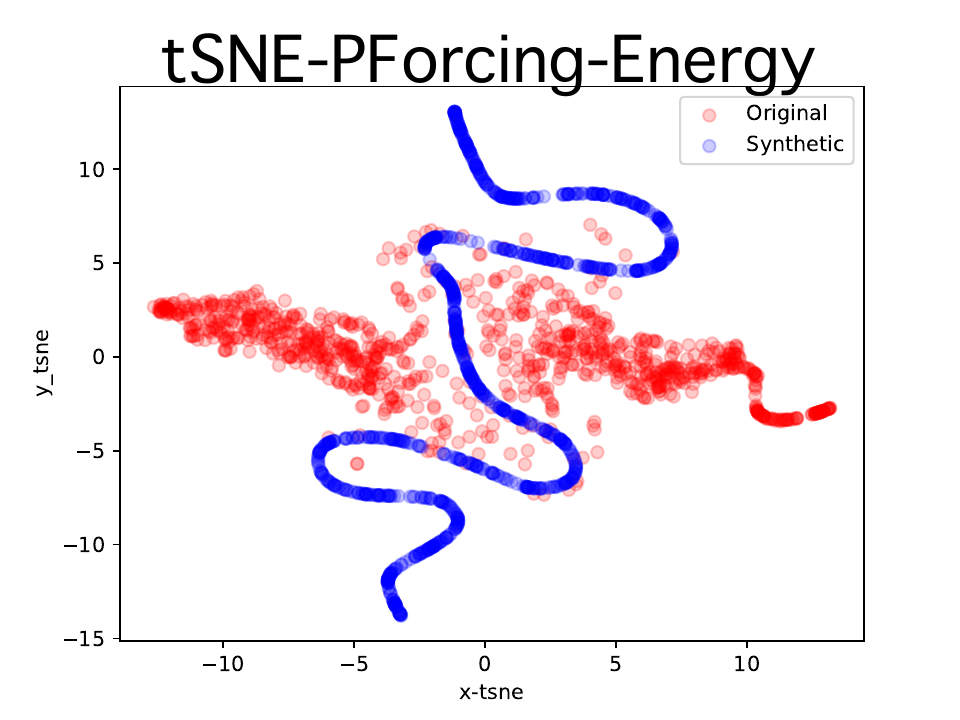}}\hfill

\caption{t-SNE plots demonstrate how well the distributions of original data samples (shown in red) align with synthetic data samples (shown in blue) generated by AAEs, GANs, T-forcing, and P-forcing methods across three different datasets.}
\label{fig:tsne_more}
\end{figure}

\begin{figure}
\centering
\subfloat{\includegraphics[width=0.163\textwidth]{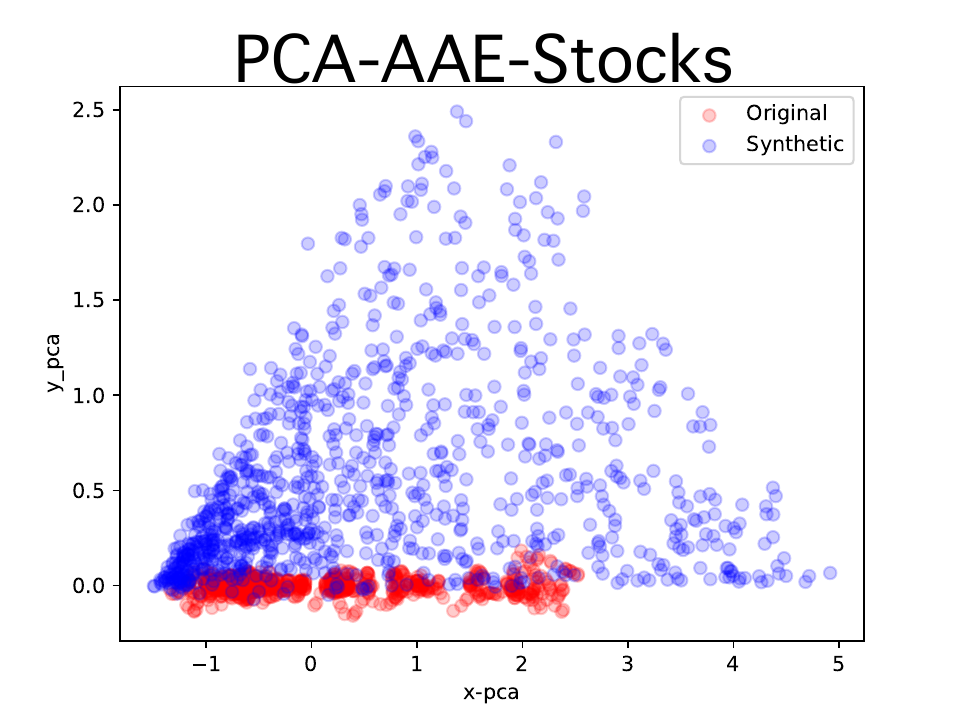}}\hfill
\subfloat{\includegraphics[width=0.163\textwidth]{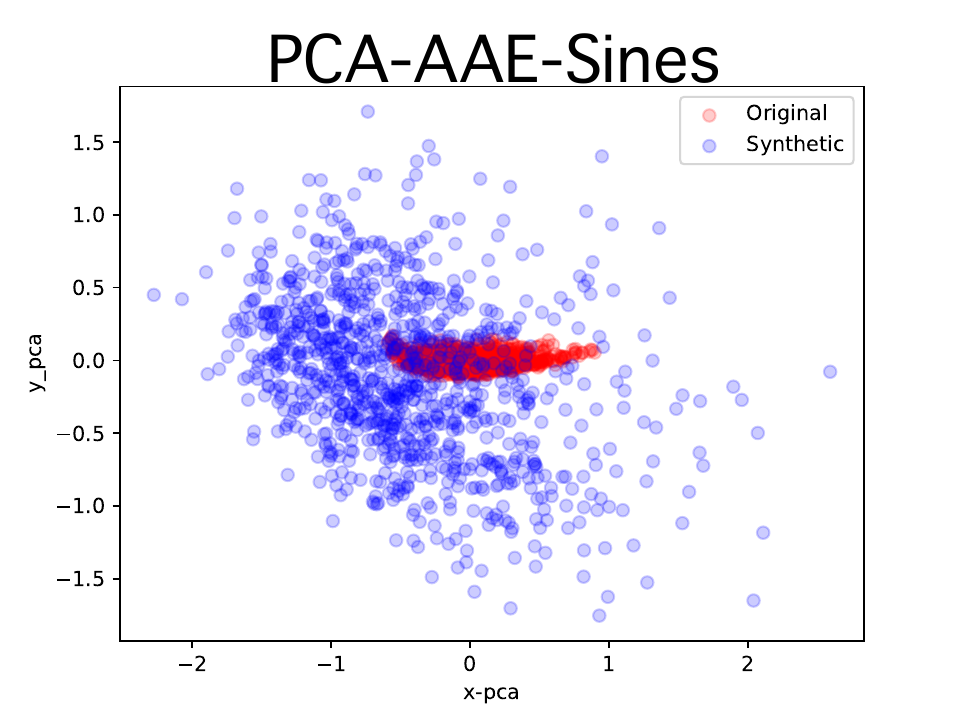}}\hfill
\subfloat{\includegraphics[width=0.163\textwidth]{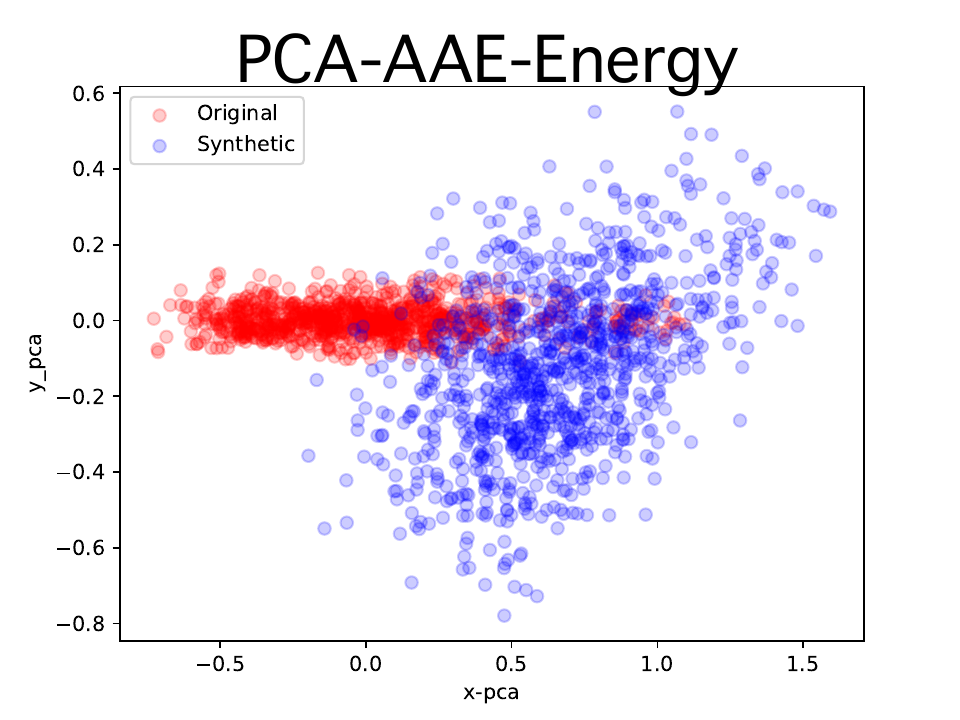}}\hfill

\vspace{-1em}

\subfloat{\includegraphics[width=0.163\textwidth]{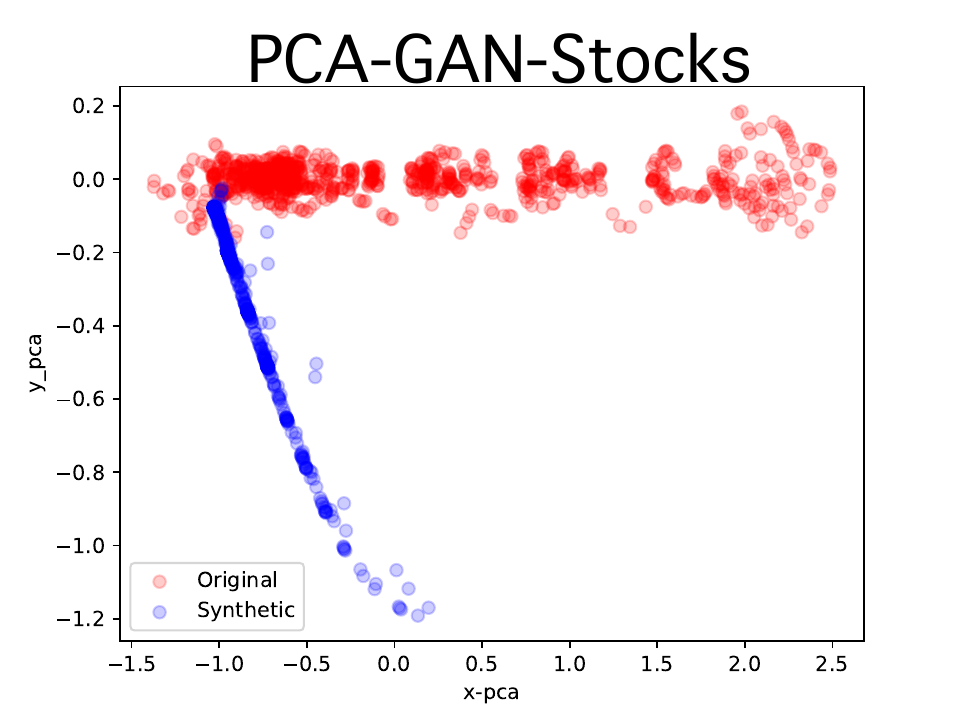}}\hfill
\subfloat{\includegraphics[width=0.163\textwidth]{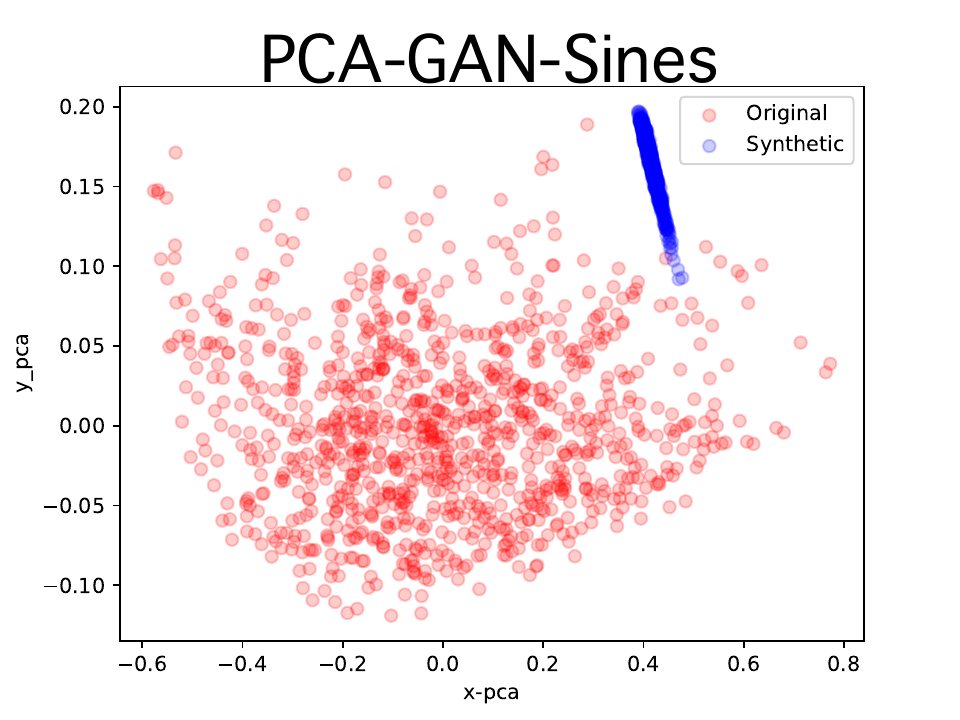}}\hfill
\subfloat{\includegraphics[width=0.163\textwidth]{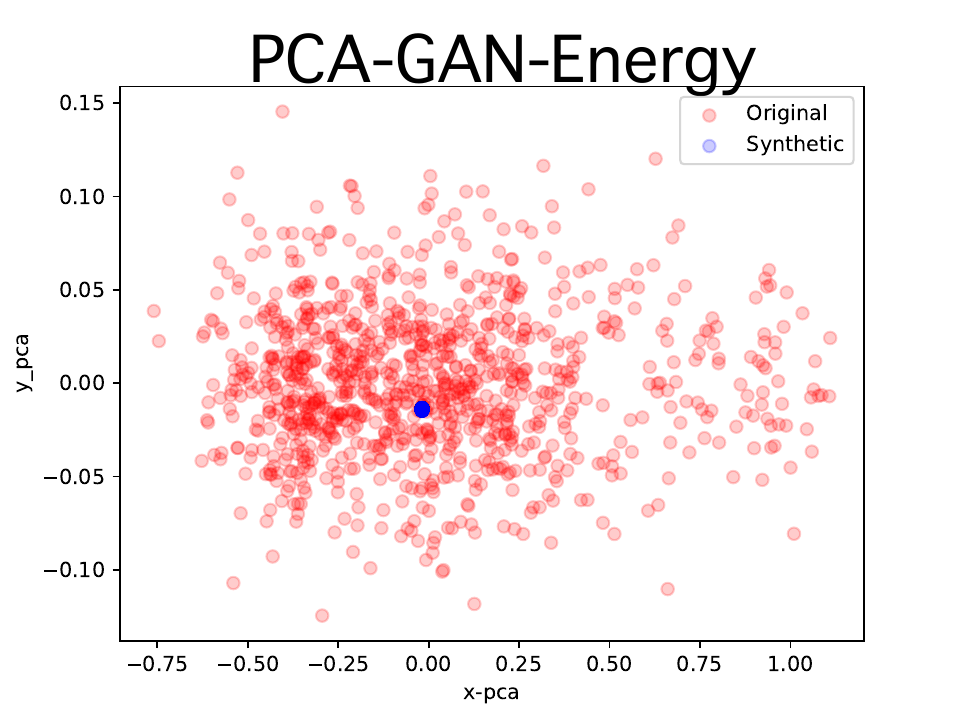}}\hfill

\vspace{-1em}

\subfloat{\includegraphics[width=0.163\textwidth]{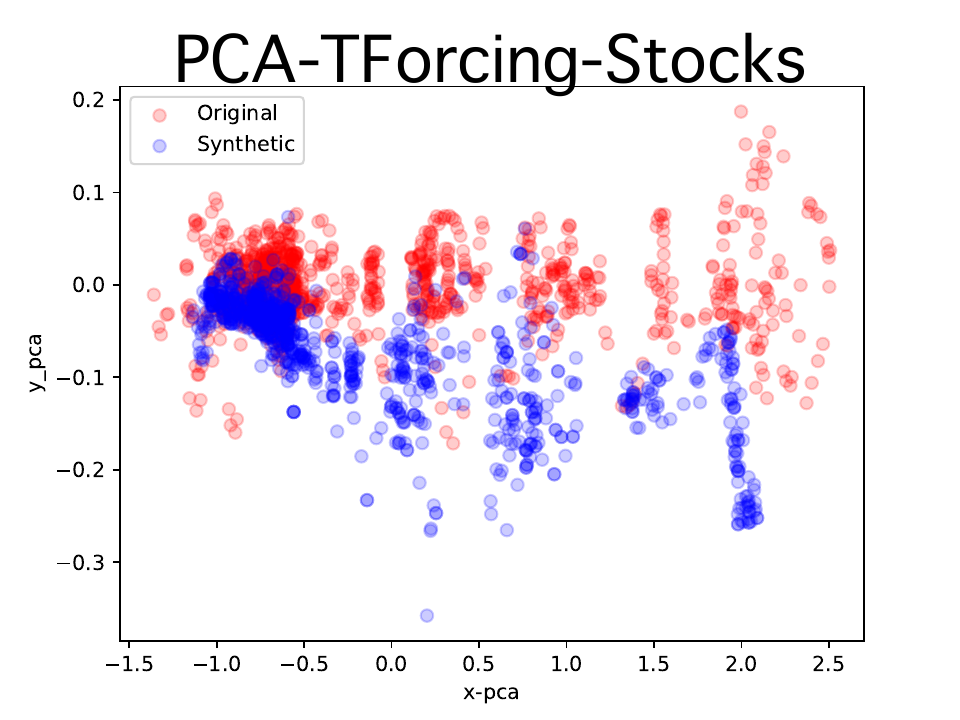}}\hfill
\subfloat{\includegraphics[width=0.163\textwidth]{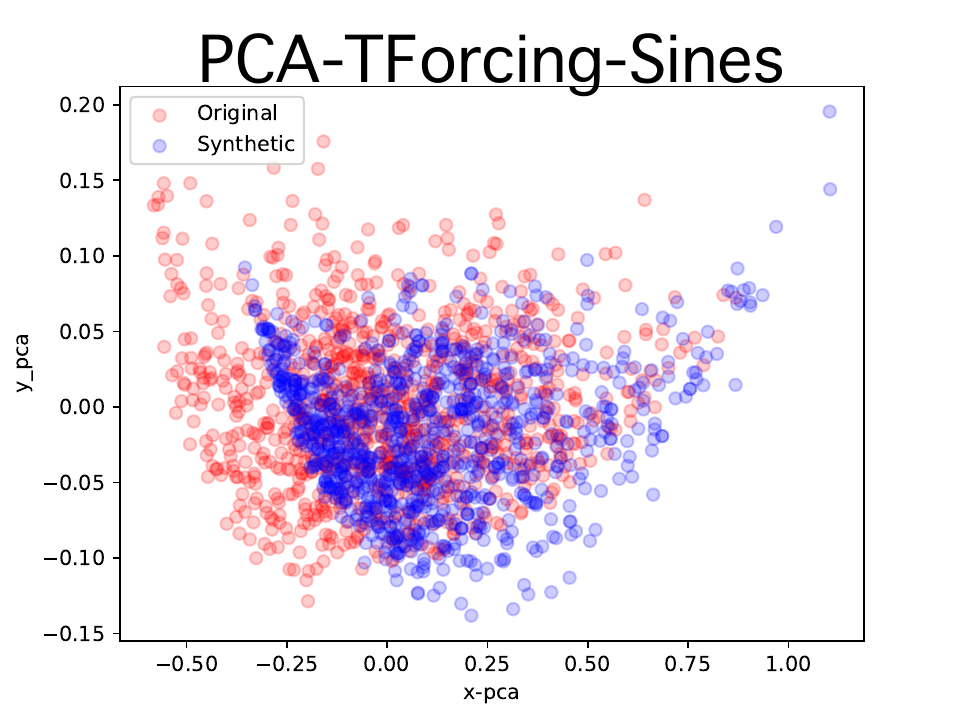}}\hfill
\subfloat{\includegraphics[width=0.163\textwidth]{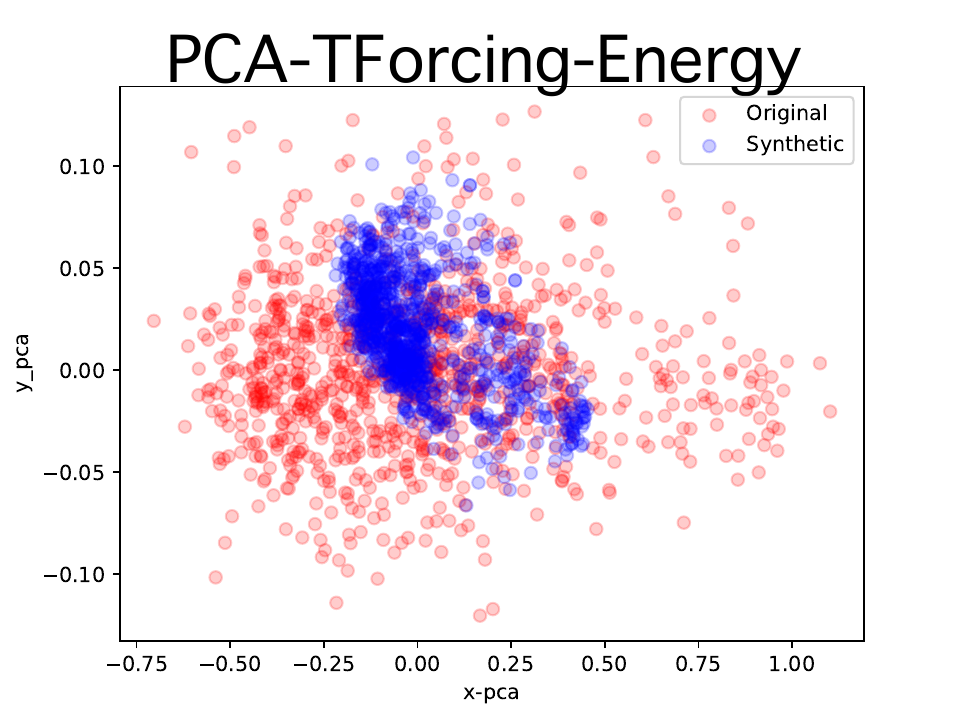}}\hfill

\vspace{-1em}

\subfloat{\includegraphics[width=0.163\textwidth]{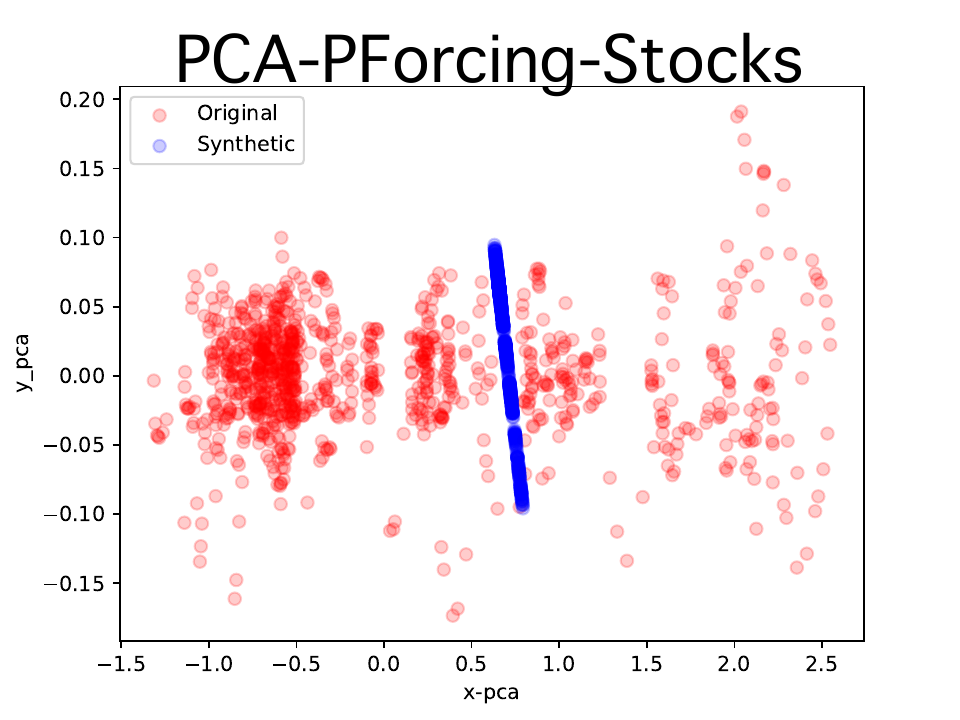}}\hfill
\subfloat{\includegraphics[width=0.163\textwidth]{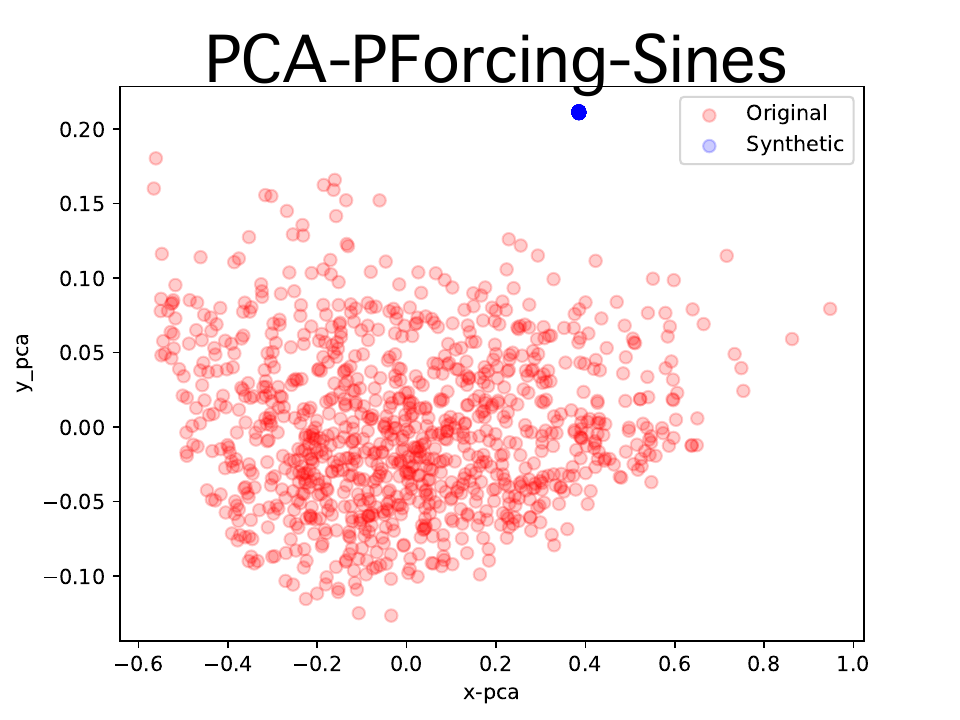}}\hfill
\subfloat{\includegraphics[width=0.163\textwidth]{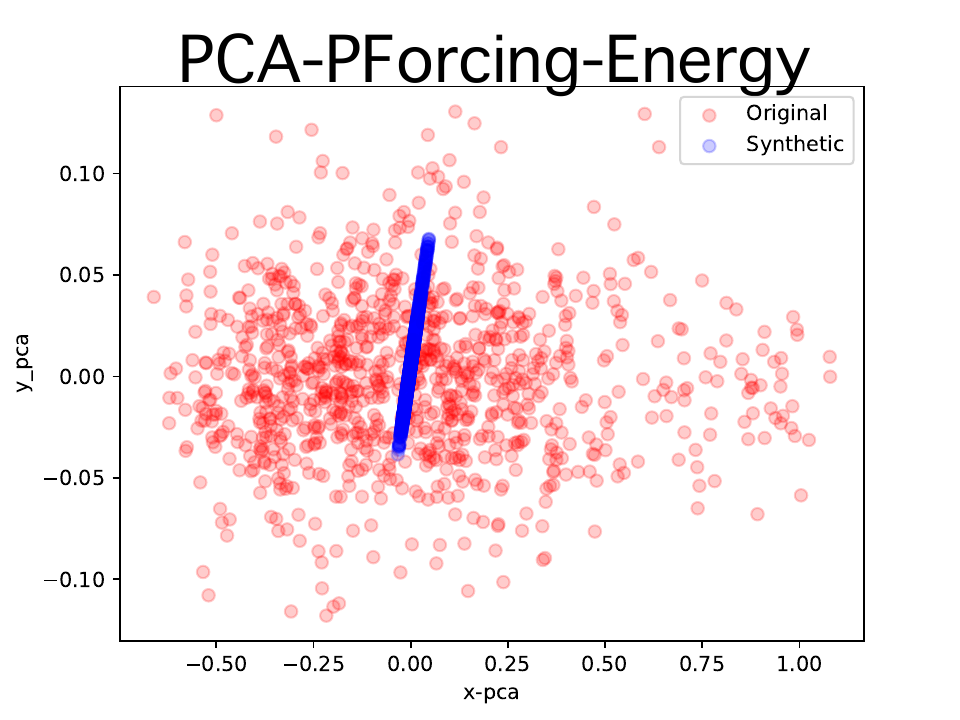}}\hfill

\caption{PCA plots illustrate the degree of alignment between the distributions of original data samples (in red) and synthetic data samples (in blue) generated by AAEs, GANs, T-forcing, and P-forcing across three datasets.}
\label{fig:pca_more}
\end{figure}

Figures \ref{fig:tsne_more} and \ref{fig:pca_more} present the t-SNE and PCA plots for the GANs, AAEs, T-forcing, and P-forcing techniques. With the exception of T-forcing, which demonstrates a stronger alignment, the other techniques struggle to effectively learn the distribution of time series data.
\label{sec:appendixA}

\section{Additional Visual Analysis.}
\label{sec:appendixB}

\begin{figure}
\centering
\includegraphics[width=0.50\textwidth]{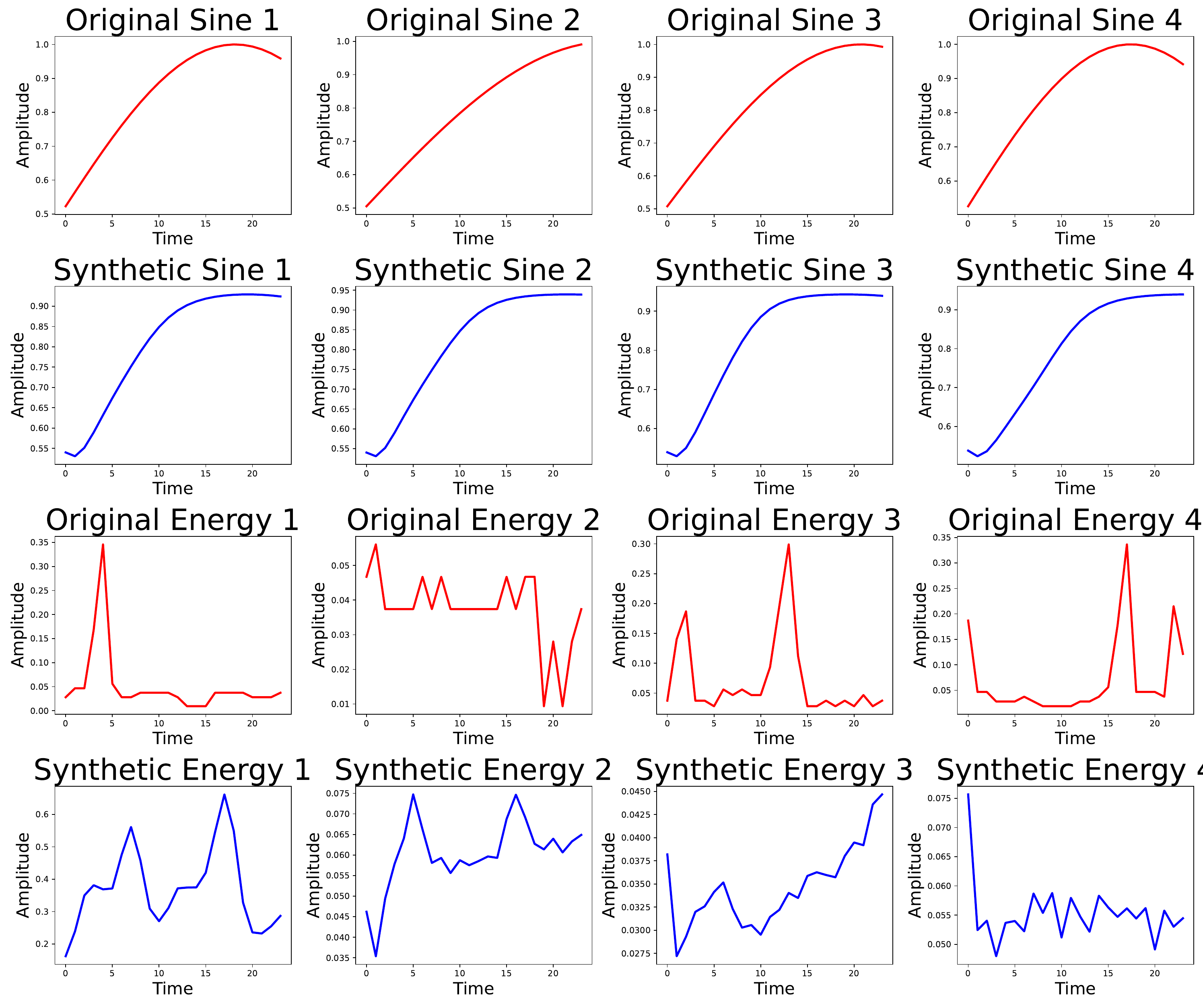}
\caption{The figure displays four randomly selected samples of both original data (red) and synthetic data (blue) generated by AVATAR for the Sines and Energy datasets.}
\label{fig:examples_sine_energy}
\end{figure}

Figure \ref{fig:examples_sine_energy} showcases both the original and synthetic data generated by AVATAR for the Sines and Energy datasets. The figure highlights AVATAR’s ability to accurately capture the smoothness of the Sines data, as well as its effectiveness in capturing the sharpness and periodicity of the Energy dataset.

\balance


\begin{thebibliography}{99}

\bibitem{Ahmadzadeh2021}
A.~Ahmadzadeh, B.~Aydin, M.~K. Georgoulis, D.~J. Kempton, S.~S. Mahajan, and R.~A. Angryk, {\em How to Train Your Flare Prediction Model: Revisiting Robust Sampling of Rare Events}, The Astrophysical Journal Supplement Series, vol. 254, no. 2, pp. 23, May 2021. [Online]. Available: http://dx.doi.org/10.3847/1538-4365/abec88.

\bibitem{EskandariNasab2024}
M.~EskandariNasab, S.~M.~Hamdi, and S.~F.~Boubrahimi, {\em Impacts of data preprocessing and sampling techniques on solar flare prediction from multivariate time series data of photospheric magnetic field parameters}, Astrophysical Journal Supplement Series, in press. [Online]. Available: http://dx.doi.org/10.3847/1538-4365/ad7c4a.

\bibitem{sfcontrastive-reza}
M.~R.~EskandariNasab, S.~M.~Hamdi, and S.~F.~Boubrahimi, {\em Enhancing Multivariate Time Series-based Solar Flare Prediction with Multifaceted Preprocessing and Contrastive Learning}, arXiv preprint, arXiv:2409.14016, 2024. [Online]. Available: https://arxiv.org/abs/2409.14016.

\bibitem{sfcontrastive-onur}
O.~Vural, S.~M.~Hamdi, and S.~F.~Boubrahimi, {\em EXCON: Extreme Instance-based Contrastive Representation Learning of Severely Imbalanced Multivariate Time Series for Solar Flare Prediction}, arXiv preprint, arXiv:2411.11249, 2024. [Online]. Available: https://arxiv.org/abs/2411.11249.

\bibitem{Abouelmehdi2017}
K.~Abouelmehdi, A.~Beni-Hssane, H.~Khaloufi, and M.~Saadi, {\em Big data security and privacy in healthcare: A Review}, Procedia Computer Science, vol. 113, pp. 73–80, 2017. [Online]. Available: https://doi.org/10.1016/j.procs.2017.08.292.

\bibitem{Cascalheira2024}
C.~J.~Cascalheira, K.~Corro, C.~Hong, T.~K.~Rohleen, O.~Trac, and M.~Beikzadeh, {\em An Analysis of Mpox Communication on Reddit vs Twitter During the 2022 Mpox Outbreak}, Sexuality Research and Social Policy, Dec. 2024. [Online]. Available: https://doi.org/10.1007/s13178-024-01058-4.

\bibitem{santosh2024}
S.~Chapagain, Y.~Zhao, T.~K.~Rohleen, S.~M.~Hamdi, S.~F.~Boubrahimi, and R.~E.~Flinn, {\em Predictive Insights into LGBTQ+ Minority Stress: A Transductive Exploration of Social Media Discourse}, 2024 IEEE 11th International Conference on Data Science and Advanced Analytics (DSAA), 2024, pp. 1--9. [Online]. Available: https://doi.org/10.1109/DSAA61799.2024.10722807.

\bibitem{Vincent2017}
E.~Vincent, S.~Watanabe, A.~A.~Nugraha, J.~Barker, and R.~Marxer, {\em An analysis of environment, microphone and data simulation mismatches in robust speech recognition}, Computer Speech \& Language, vol. 46, pp. 535–557, 2017. [Online]. Available: https://doi.org/10.1016/j.csl.2016.11.005.

\bibitem{Mikolajczyk2018}
A.~Mikołajczyk and M.~Grochowski, {\em Data augmentation for improving deep learning in image classification problem}, in Proc. 2018 International Interdisciplinary PhD Workshop (IIPhDW), 2018, pp. 117–122. [Online]. Available: https://doi.org/10.1109/IIPHDW.2018.8388338.

\bibitem{thesis24}
M.~R.~EskandariNasab, {\em Supervised Generative Adversarial Networks for Time Series Generation in Embedding Space}, Master’s Thesis, Utah State University, 2024. [Online]. Available: https://digitalcommons.usu.edu/etd2023/387.

\bibitem{Voulodimos2018}
A.~Voulodimos, N.~Doulamis, A.~Doulamis, and E.~Protopapadakis, {\em Deep Learning for Computer Vision: A Brief Review}, Computational Intelligence and Neuroscience, vol. 2018, no. 1, pp. 7068349, 2018. [Online]. Available: https://doi.org/10.1155/2018/7068349.

\bibitem{aria2022}
A.~Mehrad, E.~Nourani, and A.~Golzari Oskouei, {\em ADA-COVID: Adversarial Deep Domain Adaptation-Based Diagnosis of COVID-19 from Lung CT Scans Using Triplet Embeddings}, Computational Intelligence and Neuroscience, vol. 2022, Article ID 2564022, 17 pages, 2022. [Online]. Available: https://doi.org/10.1155/2022/2564022.

\bibitem{Li2023}
P.~Li, P.~Hosseinzadeh, O.~Bahri, S.~F.~Boubrahimi, and S.~M.~Hamdi, {\em Adversarial Attack Driven Data Augmentation for Time Series Classification}, in Proc. 2023 International Conference on Machine Learning and Applications (ICMLA), 2023, pp. 653–658. [Online]. Available: https://doi.org/10.1109/ICMLA58977.2023.00096.

\bibitem{obahri2025}
O.~Bahri, P.~Li, P.~Hosseinzadeh, S.~F.~Boubrahimi, and S.~M.~Hamdi, {\em Denoising Optimization-Based Counterfactual Explanations for Time Series Classification}, Pattern Recognition, Springer Nature Switzerland, 2025, pp. 162–179. [Online]. Available: https://doi.org/10.1007/978-3-031-78183-4\_11.

\bibitem{AlshammariTF2024}
K.~Alshammari, S.~M.~Hamdi, and S.~F.~Boubrahimi, {\em Transformer Model for Multivariate Time Series Classification: A Case Study of Solar Flare Prediction}, Pattern Recognition, Springer Nature Switzerland, 2025, pp. 238–254. [Online]. Available: https://doi.org/10.1007/978-3-031-78383-8\_16.

\bibitem{ESKANDARINASAB2024GRUCNN}
M.~EskandariNasab, Z.~Raeisi, R.~A. Lashaki, and H.~Najafi, {\em A GRU–CNN model for auditory attention detection using microstate and recurrence quantification analysis}, Scientific Reports, 14 (2024), p.~8861, doi: 10.1038/s41598-024-58886-y.

\bibitem{ar1}
S.~Bengio, O.~Vinyals, N.~Jaitly, and N.~Shazeer, {\em Scheduled sampling for sequence prediction with recurrent neural networks}, in Proc. 28th Int. Conf. Neural Information Processing Systems, 1 (2015), pp.~1171–1179.

\bibitem{ar2}
A.~Goyal, A.~Lamb, Y.~Zhang, S.~Zhang, A.~Courville, and Y.~Bengio, {\em Professor forcing: A new algorithm for training recurrent networks}, arXiv:1610.09038, 2016. [Online]. Available: https://arxiv.org/abs/1610.09038.

\bibitem{ar3}
D.~Bahdanau, P.~Brakel, K.~Xu, A.~Goyal, R.~Lowe, J.~Pineau, A.~Courville, and Y.~Bengio, {\em An actor-critic algorithm for sequence prediction}, arXiv preprint arXiv:1607.07086, 2017. [Online]. Available: https://arxiv.org/abs/1607.07086.

\bibitem{gan}
I.~J. Goodfellow, J.~Pouget-Abadie, M.~Mirza, B.~Xu, D.~Warde-Farley, S.~Ozair, A.~Courville, and Y.~Bengio, {\em Generative adversarial networks}, arXiv preprint arXiv:1406.2661, 2014.

\bibitem{gan1}
O.~Mogren, {\em C-RNN-GAN: Continuous recurrent neural networks with adversarial training}, CoRR, abs/1611.09904, 2016. [Online]. Available: http://arxiv.org/abs/1611.09904.

\bibitem{gan2}
C.~Esteban, S.~L. Hyland, and G.~Rätsch, {\em Real-valued (Medical) time series generation with recurrent conditional GANs}, arXiv e-prints, Jun. 2017, doi: 10.48550/arXiv.1706.02633.

\bibitem{aae}
A.~Makhzani, J.~Shlens, N.~Jaitly, I.~Goodfellow, and B.~Frey, {\em Adversarial Autoencoders}, arXiv:1511.05644, 2016. [Online]. Available: https://arxiv.org/abs/1511.05644.

\bibitem{Williams1989}
R.~J.~Williams and D.~Zipser, {\em A learning algorithm for continually running fully recurrent neural networks}, Neural Computation, vol. 1, no. 2, pp. 270–280, Jun. 1989. [Online]. Available: https://doi.org/10.1162/neco.1989.1.2.270.

\bibitem{Pforcing}
Y.~Ganin, E.~Ustinova, H.~Ajakan, P.~Germain, H.~Larochelle, F.~Laviolette, M.~Marchand, and V.~Lempitsky, {\em Domain-adversarial training of neural networks}, Journal of Machine Learning Research, vol. 17, no. 1, pp. 2096–2030, 2016.

\bibitem{Tforcing}
A.~Lamb, A.~Goyal, Y.~Zhang, S.~Zhang, A.~Courville, and Y.~Bengio, {\em Professor forcing: A new algorithm for training recurrent networks}, arXiv preprint arXiv:1610.09038 [stat.ML], 2016. [Online]. Available: https://arxiv.org/abs/1610.09038.

\bibitem{Brophy2023}
E.~Brophy, Z.~Wang, Q.~She, and T.~Ward, {\em Generative Adversarial Networks in Time Series: A Systematic Literature Review}, ACM Comput. Surv., vol. 55, no. 10, article 199, pp. 1–31, Oct. 2023. [Online]. Available: https://doi.org/10.1145/3559540.

\bibitem{seriesgan}
M.~R.~EskandariNasab, S.~M.~Hamdi, and S.~F.~Boubrahimi, {\em SeriesGAN: Time Series Generation via Adversarial and Autoregressive Learning}, arXiv preprint, arXiv:2410.21203, 2024. [Online]. Available: https://arxiv.org/abs/2410.21203.

\bibitem{timegan}
J.~Yoon, D.~Jarrett, and M.~van der Schaar, {\em Time-series generative adversarial networks}, in Advances in Neural Information Processing Systems, 2019. [Online]. Available: https://doi.org/10.1145/3559540.

\bibitem{variational}
D.~P.~Kingma and M.~Welling, {\em Auto-Encoding Variational Bayes}, arXiv preprint arXiv:1312.6114, 2022. [Online]. Available: https://arxiv.org/abs/1312.6114.

\bibitem{Creswell2019}
A.~Creswell and A.~A.~Bharath, {\em Denoising Adversarial Autoencoders}, IEEE Transactions on Neural Networks and Learning Systems, vol. 30, no. 4, pp. 968–984, 2019. [Online]. Available: https://doi.org/10.1109/TNNLS.2018.2852738.

\bibitem{divergence1}
M.~L.~Menéndez, J.~A.~Pardo, L.~Pardo, and M.~C.~Pardo, {\em The Jensen-Shannon divergence}, Journal of the Franklin Institute, vol. 334, no. 2, pp. 307–318, 1997. [Online]. Available: https://doi.org/10.1016/S0016-0032(96)00063-4.

\bibitem{divergence2}
J.~Shlens, {\em Notes on Kullback-Leibler Divergence and Likelihood}, arXiv preprint arXiv:1404.2000, 2014. [Online]. Available: https://arxiv.org/abs/1404.2000.

\bibitem{UCI_Appliances_Energy}
L.~M.~Ibarra Candanedo, V.~Feldheim, and D.~Deramaix, {\em Data driven prediction models of energy use of appliances in a low-energy house}, Energy and Buildings, vol. 140, pp. 81–97, 2017. [Online]. Available: https://doi.org/10.1016/j.enbuild.2017.01.083.


\bibitem{ganevaluation1}
L.~Theis, A.~van den Oord, and M.~Bethge, {\em A note on the evaluation of generative models}, arXiv preprint arXiv:1511.01844, 2016. [Online]. Available: https://arxiv.org/abs/1511.01844.

\bibitem{ganevaluation2}
Y.~Wu, Y.~Burda, R.~Salakhutdinov, and R.~Grosse, {\em On the Quantitative Analysis of Decoder-Based Generative Models}, arXiv preprint arXiv:1611.04273, 2017. [Online]. Available: https://arxiv.org/abs/1611.04273.

\bibitem{tsne}
L.~van der Maaten and G.~Hinton, {\em Visualizing data using t-SNE}, Journal of Machine Learning Research, 9 (2008), pp.~2579–2605. [Online]. Available: http://jmlr.org/papers/v9/vandermaaten08a.html.

\bibitem{pca}
F.~B. Bryant and P.~R. Yarnold, {\em Principal-components analysis and exploratory and confirmatory factor analysis}, in Reading and Understanding Multivariate Statistics, L.~G. Grimm and P.~R. Yarnold, Eds., American Psychological Association, Washington, DC, 1995, pp.~99–136.


\end{thebibliography}
\end{document}